%% file: acl_latex.tex
\documentclass[11pt]{article}

\usepackage[final]{acl}
\usepackage{times}
\usepackage{latexsym}

\usepackage{graphicx}
\usepackage{enumitem}
\usepackage{booktabs}
\usepackage{xurl}
\usepackage{hyperref}
\usepackage{url}
\usepackage{multirow}
\usepackage{amsmath}
\usepackage{amssymb}
\usepackage{float}
\usepackage{wrapfig}
\usepackage[table]{xcolor}
\usepackage{algorithm}
\usepackage{algorithmic}
\usepackage{todonotes}
\usepackage[T1]{fontenc}
\usepackage[utf8]{inputenc}
\usepackage{microtype}
\usepackage{inconsolata}
\usepackage{graphicx}
\usepackage[most]{tcolorbox}
\tcbuselibrary{skins}
\usepackage{listings}

\title{CorporateBench: Large-Scale Q\&A Benchmarking\\ with Temporal Knowledge Bases}

\author{Sil Hamilton\textsuperscript{1, 2}\thanks{Correspondence to: \href{mailto:srh255@cornell.edu}{srh255@cornell.edu}}, Albert Yu Sun\textsuperscript{1}, Oscar J. Romero\textsuperscript{1}, Carl-Leander Henneking\textsuperscript{1}, \\ 
\textbf{David Mimno\textsuperscript{2}}, \textbf{Bishan Yang\textsuperscript{1}, Igor Labutov\textsuperscript{1}}  \\
\textsuperscript{1} Epiq AI Labs \textsuperscript{2} Cornell University \\
}

\begin{document}
\maketitle
\begin{abstract}
\input{writing/abstract}
\end{abstract}

\section{Introduction}
\input{writing/introduction}

\section{Related Work}
\input{writing/related_work}

\section{Dataset Construction}
\input{writing/benchmark_construction}

\section{Dataset Validation}
\label{sec:validation}
\input{writing/validation}

\section{Tasks}

\input{writing/benchmark}

\section{Results}
\input{writing/results_new}

\section{Conclusion}
\input{writing/conclusion}

\section*{Limitations}

We acknowledge the following limitations:

\paragraph{Scope limited to communication networks.}
\textsc{CorporateBench} focuses on email, which remains the dominant document type in corporate communication networks: the Enron corpus, the largest publicly available corporate dataset, is composed almost entirely of email \citep{klimt2004enron}.
Real corporate settings additionally involve communication channels such as Slack and Microsoft Teams, and document types such as technical specifications, financial reports, and presentations, each with distinct structural characteristics.
Extending \textsc{CorporateBench} to additional communication channels is a direction for future work.

\paragraph{Simplified document types.}
Our benchmark focuses primarily on email communications. 
Real corporate settings involve newer communication channels such as Slack and Microsoft Teams, and documents themselves may be diverse across document types such as technical specifications, financial reports, spreadsheets, presentations.
Each document type has distinct structural characteristics that may present different challenges for LLMs beyond just emails.
We note that emails still remain a large majority document type in recent corporate discovery contexts \citep{gallivan2024ediscovery}.

\paragraph{Temporal dynamics and evolution.}
In simulating one quarter of company activity, we explicitly do not model longer-term organizational evolution such as strategic pivots, mergers and acquisitions, cultural shifts, or the gradual accumulation of institutional knowledge over years. 
We leave it for future work to expand beyond the 90-day simulation window to surface longer term phenomena.

\paragraph{LLM-generated artifacts.}
Despite efforts to ensure diversity through topics, personalities, and formality levels, the documents are generated by Claude Haiku 4.5, which may introduce systematic biases or artifacts in writing style, vocabulary, and document structure. 
Models evaluated on this benchmark may perform differently on human-written corporate documents.
We see our tasks as a ``lower bound'' in terms of difficulty, where LLM agents need to be able to solve our tasks first before solving likely messier, more difficult, and more complex human-written corpora.

\paragraph{QA datasets.}
There are a few assumptions and/or design decisions that affect the QA datasets. These are offered to the reader in \autoref{app:qa-dataset_assumptions}.

\section*{Ethical Considerations}
\input{writing/ethics}

\bibliography{custom}

\appendix

\input{writing/appendix}

\end{document}

%% file: writing/abstract.tex
LLMs are increasingly able to answer complex questions about enterprise-scale document collections.
But evaluation is hard: companies don't want to share internal communications, and synthetic datasets have been overly simple.
We present \textsc{CorporateBench} (\textsc{CB}), a human-validated multi-task Q\&A benchmark whose scale approaches the conditions LLMs encounter in corporate communication networks, with evaluation corpora surpassing 230{,}000 documents.
\textsc{CB} evaluates LLMs across two dimensions (information extraction and knowledge base querying) through four synthetically generated firms ranging from 12 to 10{,}000 employees.
Each corpus is sampled from a temporally evolving knowledge base describing a consistent world, guaranteeing cross-document logical consistency even across hundreds of thousands of documents.
We evaluate five LLMs on \textsc{CB}, revealing increasingly poor performance as input size approaches realistic scales.
\textsc{CB} provides LLM developers a metric for corporate communication reasoning, filling a crucial gap in the benchmarking ecosystem.

%% file: writing/introduction.tex
\begin{figure}[t]
    \centering
    \includegraphics[width=0.485\textwidth]{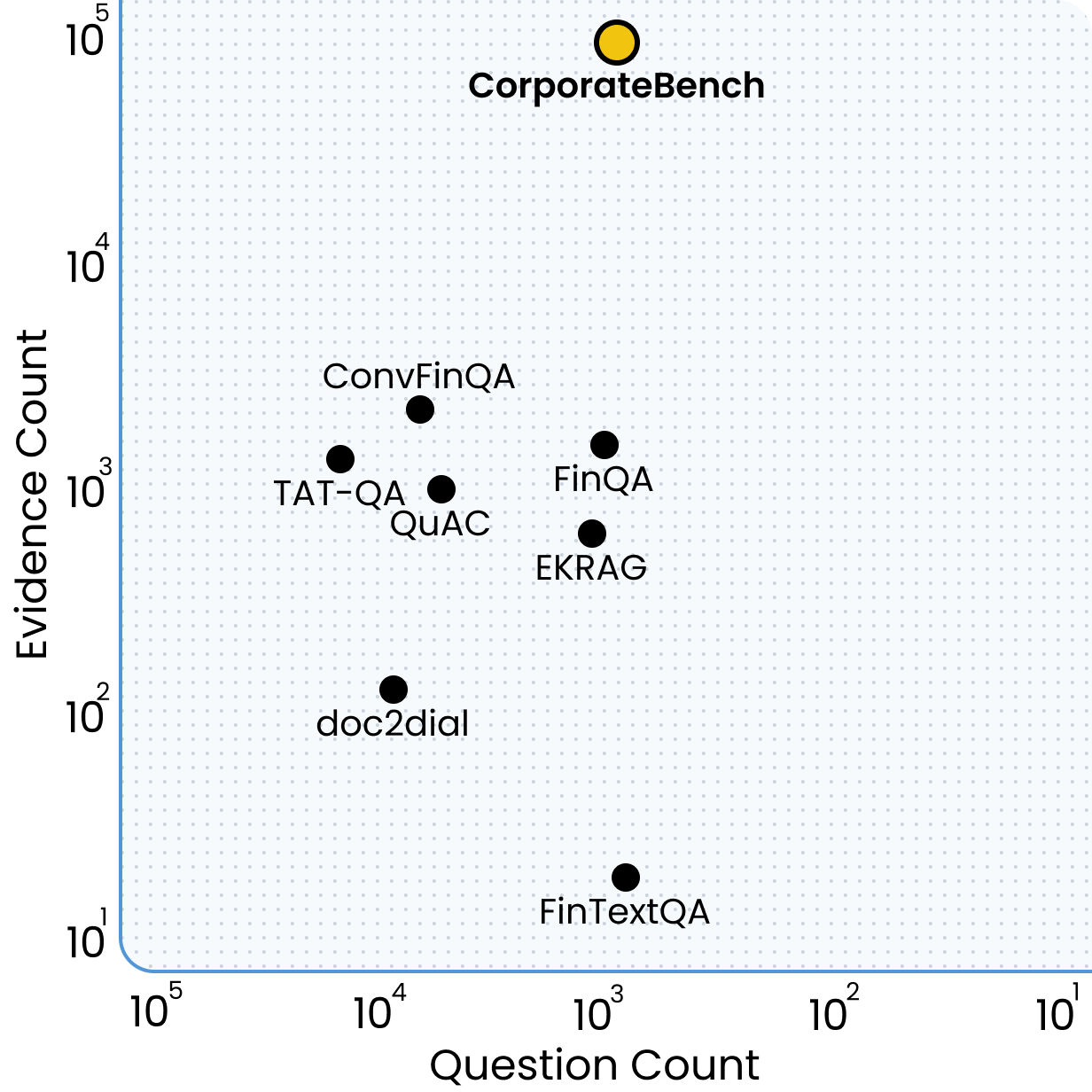}
    \vspace{-\baselineskip}
    \caption{\textsc{CorporateBench} advances the state of the art in multi-document QA benchmarking when compared against nine previous benchmarks in evidence and question counts. \textsc{CB} achieves a high ratio of \textbf{87.6} documents per question, with the closest benchmark (EKRAG) achieving a ratio of 1.5 \citep{yuEKRAGBenchmarkRAG2025}. 
    A high ratio indicates that questions require synthesizing information across many documents, which better reflects the complexity of real enterprise knowledge work.}
    \label{fig:benchmark-comparison}
\end{figure}

Complex organizations like corporations produce massive quantities of internal documents through employees communicating and working together.
LLMs are increasingly helpful for knowledge work, document analysis, and other tasks requiring general reasoning about organizations and their social networks \citep{Raza_Jahangir_Riaz_Saeed_Sattar_2025, brachman2025currentfutureuselarge}.
But despite significant advances in extending large language model (LLM) context windows \citep{survey_long_context, longformer}, state-of-the-art LLMs continue to struggle with complex reasoning over many documents as regularly required in enterprise scenarios \citep{liuLostMiddleHow2024, hsiehRULERWhatsReal2024a}. 
This deficiency hinders wide-scale LLM adoption in industry, but current evaluation metrics fail to capture this failure mode.

Common long context understanding tests do not reproduce the complexity and interconnectedness of real corporate environments.
Synthetic benchmarks \citep{kamradtNeedleHaystackPressure2023} are excessively simplified, while more realistic benchmarks based on real-world enterprise data are scarce due to non-disclosure agreements \citep{zhangEnterpriseBenchmarksLarge2024}.
The few existing enterprise benchmarks moreover fail to adequately reproduce corporate conditions due to limitations in scale and realism.
Enterprise corpora can scale to millions of documents, making the cost of generating a whole corpus from scratch a major concern when designing benchmarks.

To address these limitations, we introduce \textsc{CorporateBench}, an enterprise benchmark containing large and internally consistent  corporate communication networks with verifiable ground truth as shown in \autoref{fig:fig1}.\footnote{We release data and code at \url{https://huggingface.co/datasets/epiq-ai-labs/corporatebench}.}
\textsc{CorporateBench} bridges the synthetic-real gap by first sampling structures, and then documents, from procedurally generated knowledge bases (KBs) that capture complex relationships between employees, teams, projects, and tasks over time. 
These KBs serve as our foundation for generating diverse corporate corpora at realistic scales. Our primary contributions are as follows:

\begin{itemize}[itemsep=-5pt]
    \item \textbf{Comprehensive company simulation.} We adapt KB-to-corpus generation to produce corporate document collections closely mimicking the real world, preserving temporal sequencing and role-respecting communication directions across teams and departments.
    \item \textbf{Deeply interconnected evidence sets.} We construct a large-scale benchmark in which answering typical queries requires integrating evidence across many documents, stressing multi-document retrieval and reasoning over enterprise-scale communication sets whose total size can exceed context windows.
    \item \textbf{Enterprise-scale tasks.} We provide a task suite with deterministic labels directly computed from the KB, enabling reproducible evaluation. This KB-grounded design targets the gap between models’ extended context windows and genuine long context understanding in enterprise settings.
\end{itemize}

%% file: writing/related_work.tex
\paragraph{Benchmarking LLMs in long contexts.}
Enterprise datasets are naturally long context.
Contemporary LLMs can now attend to millions of tokens \citep{openaiGPT4oSystemCard2024,grattafioriLlama3Herd2024,teamGemini15Unlocking2024}, but these models suffer from poor long context understanding \citep{kaplanScalingLawsNeural2020,liuLostMiddleHow2024}.
One method for identifying these failure modes is the ``needle-in-a-haystack'' test and its descendants \citep{kamradtNeedleHaystackPressure2023,vodrahalliMichelangeloLongContext2024a,openaiOpenAIMRCRLong2025}, but these synthetic benchmarks trade realism for generation speed, leading to newer benchmarks testing models over documents like books and movie scripts \citep{kociskyNarrativeQAReadingComprehension2018,anLEvalInstitutingStandardized2023a,karpinskaOneThousandOne2024,hamilton2025too} whose limited number fail to qualify as enterprise scale.
Researchers seeking solutions have introduced enterprise benchmarks based on financial records \citep{deusserKPIEDGARNovelDataset2022,xieFinBenHolisticFinancial2024,xuSuperCLUEFinGradedFineGrained2024}, legal \citep{manorPlainEnglishSummarization2019,guhaLegalbenchCollaborativelyBuilt2023,feiLawBenchBenchmarkingLegal2023,ryanEnronQAPersonalizedRAG2025}, knowledge work \citep{boisvertWorkArenaCompositionalPlanning2024,drouinWorkArenaHowCapable2024,stylesWorkBenchBenchmarkDataset2024,yaotbenchBenchmarkToolAgentUser2024,xuTheAgentCompanyBenchmarkingLLM2025}, and other general corporate tasks \citep{jiangEnhancingQuestionAnswering2024,zhangEnterpriseBenchmarksLarge2024,wangEnterpriseLargeLanguage2025}.
These benchmarks suffer from low quality data sources, typically transforming non-corporate data to avoid violating non-disclosure agreements.
An alternative approach involves generating diverse synthetic enterprise corpora with LLMs emulating knowledge work \citep{choubeyBenchmarkingDeepSearch2025,huangCRMArenaUnderstandingCapacity2025,vishwakarma-etal-2025-llms}, but existing implementations suffer from small scale (ca. $\leq$30,000 documents) and problematic ground truth because datasets are created via prompting LLMs and there is no single method for guaranteeing LLM output will match all desired characteristics.
Matching real corporate scale therefore demands new generation methods.

\paragraph{Synthetic knowledge bases.}
Because enterprise environments are structurally self-similar (e.g. employees exist at all levels of an organization), they can be conveniently described in a knowledge base.
KBs have long been used to represent human knowledge \citep{bollackerFreebaseCollaborativelyCreated2008,talmorWebKnowledgeBaseAnswering2018}, and more recently for producing grounded synthetic documents \citep{agarwal2021knowledge}, but complex systems like enterprise environments were rarely profiled in early studies due to difficulties with parsing relations from unstructured text \citep{kwiatkowskiScalingSemanticParsers2013,yangEmbeddingEntitiesRelations2015,yangLeveragingKnowledgeBases2019}.
LLMs have since proven adept at this task \citep{labutovMultiRelationalQuestionAnswering2019,sunThinkonGraphDeepResponsible2023,machadoLLMStoreLeveraging2024,sahayAutoKBAutomatedCreation2025,gongPhantomWikiOnDemandDatasets2025a} and likewise converting text to structured queries \citep{labutovLearningLearnSemantic2019,trivediMuSiQueMultihopQuestions2022,wangKnowledGPTEnhancingLarge2023,sunDocs2KGHumanLLMCollaborative2025,chenOneEvalBenchmarkingLLM2025}.
We look to KBs to provide a single, arbitrarily detailed world state from which we can (i) synthesize document collections at realistic, million-scale sizes and (ii) derive exact ground truth by querying the underlying graph.

%% file: writing/benchmark_construction.tex
\label{sec:construction}

\begin{figure*}[t!]
    \vspace{-0.1\baselineskip}
    \centering
    \includegraphics[width=\textwidth]{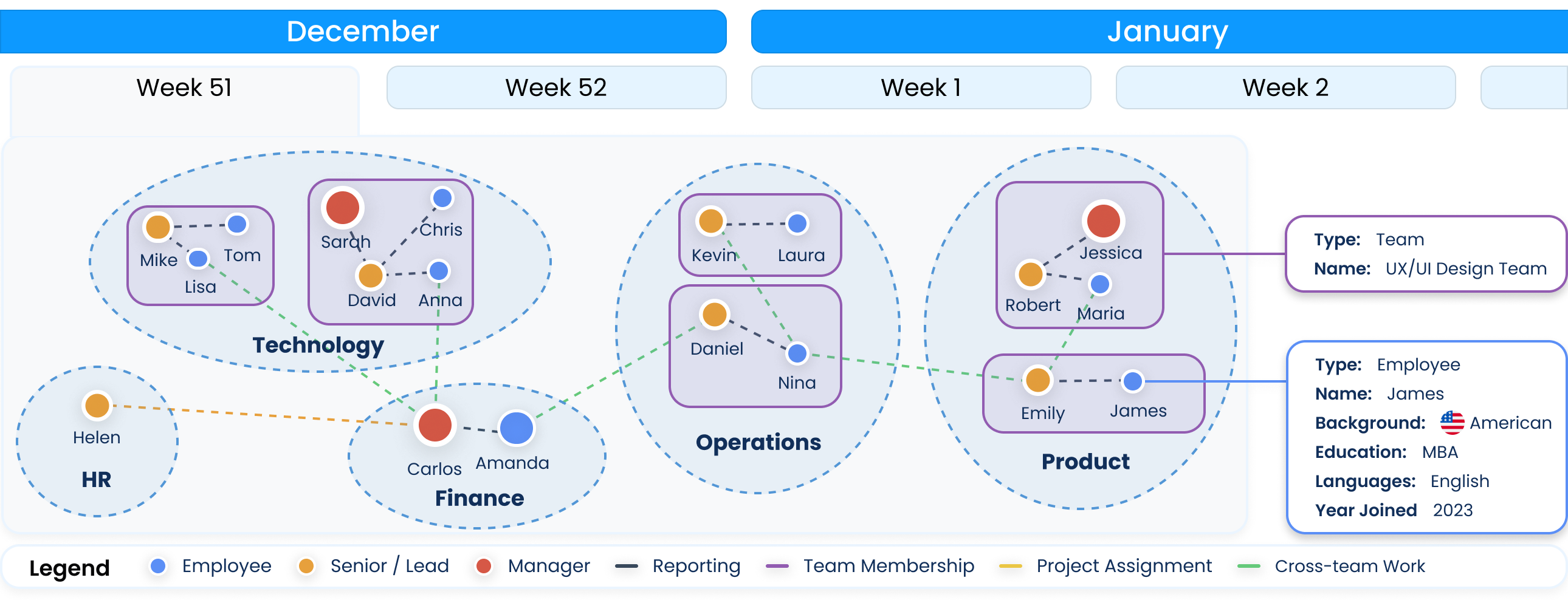}
    \vspace{-\baselineskip}
    \caption{\textsc{CorporateBench} contains four synthetic companies spanning different industries, headcounts, and business models. The figure shows a vertical slice of a hypothetical company and its social network at a given time.}
    \label{fig:fig1}
\end{figure*}

We begin constructing our synthetic enterprise benchmark by establishing ground truth.
In this section we describe a corpus construction pipeline whose output enjoys two properties:

\begin{itemize}
    \item \textbf{Arbitrary scale.} For any desired corpus size $n$, the pipeline can generate $\geq n$ documents while maintaining consistent quality across all generated documents, maintaining future relevance as LLM context windows grow.
    \item \textbf{Logical consistency.} $\forall d \in C$, no statement in document $d$ contradicts any fact established in our ground truth knowledge base.
\end{itemize}

We first generate a knowledge base representing the world of a company $C$, encoding all entities together with their properties and relationships.
We then generate emails that substantiate organizational relationships, tasks, and project progress from the KB, enforcing consistency by inserting ``evidence'' strings revealing relationships.
This pipeline satisfies logical consistency because under ideal circumstances the graph can be reconstructed from all documents sampled from it, and arbitrary scale because documents are templated.

\subsection{Defining the Knowledge Base}
We begin with a formal ontology capturing the desired semantics of our companies, written in Terse RDF Triple Language (Turtle; \citealt{beckett2014rdf}).
We formally define a company $C = (D, T, E)$ as the union of all employees $E$ organized by teams $T$ and departments $D$, where each team strictly belongs to one department and all employees strictly belong to one team, barring department leads and executives.
Serializing our companies with an OWL parser guarantees consistency.

Our ontology defines two base classes, \texttt{Entity} and \texttt{Relationship}, serving as nodes and edges in the knowledge base.
We extend these with organizational entities (companies, departments, teams, employees) and work matter (projects, meetings, tasks), linked by seven relationship predicates: \texttt{MemberOf}, \texttt{ReportsTo}, \texttt{WorksOn}, \texttt{WorksAt}, \texttt{Attends}, \texttt{Organize}, and \texttt{BelongsTo}.
These tie together employees along three axes: organizational (membership and reporting hierarchies), work-wise (task assignments), and communication (meeting attendance and organization).

\subsection{Generating Companies}
\label{par:ontology}
To generate company knowledge bases, we run the following procedure conditioned on the type of company and desired number of employees:

    \paragraph{1. \textbf{Produce company hierarchy.}}
    We logarithmically scale department and team counts with the desired employee count, targeting an average team headcount of 5--10 and department headcount of 8--68 when scaling from $10^1$ to $10^4$ employees.\footnote{We select these values from real-world distributions.}
    C-suite scales from 1 to 13 executives; departments and teams are equipped with one or more managers.
    We link all entities with \texttt{ReportsTo} and \texttt{MemberOf} relationships and verify the hierarchy forms a valid tree with no orphan nodes.
    
    \paragraph{2. \textbf{Simulate work over time.}}
    We simulate the company over the course of one quarter, or 90 days.
    For each day, we randomly assign each employee a task with $P = 1/45$ such that each employee completes between 1 to 3 tasks by the end of the quarter.\footnote{A more realistic workload would have employees complete 1 to 3 tasks a day. We purposefully limit this value to ensure uniform relationship counts (cf. \autoref{tab:company_datasets}).}
    We track task assignment and completion dates.
    We verify that all task assignments fall within the simulated quarter and that no employee exceeds the target of 1 to 3 tasks.

    \paragraph{3. \textbf{Simulate meetings over time.}}
    We simulate meetings for employees across all six meeting types: direct reports, team meetings, task collaboration sessions, executive meetings, project reviews, and department meetings. 
    The meeting generation process considers employee roles, reporting structures, team memberships, and project assignments to determine appropriate meeting schedules. 
    \autoref{app:meetings} includes more details on meetings.
    We confirm all generated meetings respect role constraints, e.g. that direct report meetings only pair employees with their managers.
    
    \paragraph{4. \textbf{Update company hierarchy.}}
    At the end of the quarter we update the company hierarchy by hiring an additional $\approx$10\% employees into a random assortment of teams.
    This process necessarily forces the splitting of those teams to maintain the desired average headcount across all teams, producing additional \texttt{MemberOf} relationships.
    This process ensures that not all \texttt{MemberOf} relationships are valid at the end of the quarter, thus requiring at least two documents to fully ``reveal'' employment status when sampling evidence.
    We verify that new hires are correctly linked with fresh \texttt{MemberOf} and \texttt{ReportsTo} relationships and that prior relationships are marked with appropriate end dates.
    
    \paragraph{5. \textbf{Annotate all entities.}}
    The company hierarchy at this point contains many entities, but no semantic properties such as names or titles.
    We prepare a one-paragraph ``story'' for each company in \autoref{fig:fig1} containing details such as company origin, motto, focus, industry, etc.\footnote{We provide all stories in \autoref{app:worlds}.}
    We then recursively pass each entity beginning at the top of the hierarchy down to the bottom to Claude Haiku 4.5 prompting it to produce a set of appropriate properties for each entity given its type and size.
    We finally produce aliases for employees and projects for diversity.
    We manually inspect a sample of annotated entities to confirm properties are contextually appropriate.
    
    \paragraph{6. \textbf{Produce triples.}}
    We finally render all entities, properties, and relationships in a sequence of triples as described in \autoref{par:ontology}.
    We store the graph in a N-Quads file \citep{tomaszuk2020rdf}, compressing it with the ontology.
    We validate the serialized graph against the OWL ontology to ensure no constraint violations.

\newtcolorbox{simplegray}[2][]{  colback=gray!10,
  colbacktitle=gray!25,
  coltitle=black,
  top=1pt,
  bottom=1pt,
  colframe=gray!40,
  fonttitle=\bfseries,
  title=#2,
  enhanced,
  attach boxed title to top center={yshift=-2mm},
  minipage boxed title*=-4mm,
  halign title=center,    boxed title style={    sharp corners,
    boxrule=0pt,
    colframe=gray!40,
  },
  sharp corners,
  boxrule=0.5pt,
  before upper={#1},
  after skip=1.5em,
}

\lstset{
  basicstyle=\small\ttfamily,
  breaklines=true,
  breakatwhitespace=true,
  columns=flexible,
  keepspaces=true,
  breakindent=0pt,
  resetmargins=true,
}

\subsection{Generating Documents}
\label{subsec:documents}

\begin{figure}[t!]
\begin{simplegray}{}
\begin{lstlisting}[escapechar=§]
From: §\textbf{Emilia A.}§ <eanglada@hotmail.com>
To: §\textbf{Jared C.}§ <jaredchartier@gmail.com>
Date: §\textbf{2024-01-10}§
Subject: Addressing code quality concerns in our operations

Hi §\textbf{Jared}§, I've been reflecting on how our current business operations could benefit from a more intentional approach to §\textbf{managing technical debt}§. As our codebase continues to grow, I'm noticing that we're accumulating layers of complexity that might slow down our team's ability to iterate efficiently. I think it would be valuable for us to establish a clearer strategy around this.

Building on that concern, §\textbf{Jared, I wanted to get your direction here}§ on how we should prioritize which areas to tackle first. I'm thinking we could start by identifying the components that have the highest maintenance burden and then work through a systematic refactoring plan. Would you be open to discussing this further so we can align on the best path forward for our technical infrastructure?

Best,
§\textbf{Emilia Anglada}§
\end{lstlisting}
\end{simplegray}
\caption{An email generated for \textsc{CorporateBench}. Employees, topic, and evidence strings are in bold.}
\label{fig:email_example}
\end{figure}

We generate emails that surface a set of one or more relationships $r \in R$ in our KB, where an email contains a sender, recipient, subject, body, and timestamp as retrieved from the KB. 
We guarantee we substantiate relationship triples by adding \textbf{evidence string(s)} that explicitly reveal the relationship $r$ from a predefined set of 3,217 strings across all seven \texttt{Relationship} types, split into 360 strings per temporal bound and 40 strings per level of difficulty.\footnote{For instance, "Just going through the {{object}} onboarding materials today" is an evidence string that substantiates that an employee's WorksOn relationship has begun.}
Each string differs in how explicit the string is in revealing the relationship, and the time frame it reveals (either beginning or end of a relationship).
Meetings use a separate set of evidence strings, illustrating task and project progress by the meeting date, with strings binned by progress up to 25\%, 50\%, 75\%, and 100\%. 
We provide examples in \autoref{app:evidence-strings}. We finish the email by passing the evidence string(s) to Claude Haiku 4.5 to in-fill the remainder of the email. 
An example is given in \autoref{fig:email_example}.
We use the following strategies to enhance lexical diversity in the output:

\paragraph{\textbf{Format.}}
    Our email formatting conventions (headers, signatures, reply structure) are modeled on the Enron corpus \citep{klimt2004enron}, a publicly available corporate communication dataset, to ensure structural realism.

    \paragraph{\textbf{Topic.}}
    To ensure diversity of synthetically generated data at scale, we sample a random topic $\mathcal{T}$ and use it to generate the majority of each email.
    Topics are sampled from a Dirichlet distribution with $\alpha=0.75$ to mimic real-world distributions where topic frequency is non-uniform.
    There are two sets: 3,000 business topics (e.g. clinical trials), and 3,000 non-business topics (e.g. food, travel, and transportation).
    Meeting topics are sampled from a separate set of 60 topics.\footnote{We provide more details in \autoref{app:topic-levels}.}
    
    \paragraph{\textbf{Personality and level of formality.}}
    We assign each employee one of sixteen personalities sourced from the Myers–Briggs Type Indicator typology \citep{briggs1976myers}.
    At the document level, we instruct the model to write emails in either a formal or a casual manner.
    Qualitative observations indicate these contribute positively to lexical diversity.

%% file: writing/validation.tex
\renewcommand{\arraystretch}{0.7}
\begin{table*}[t!]
\centering
\small
\begin{tabular*}{0.95\textwidth}{@{}l*{4}{>{\raggedleft\arraybackslash}p{0.16\textwidth}}@{}} 
\toprule
\textbf{Company} & \textbf{Zenith} & \textbf{Streamvibe} & \textbf{Biocure} & \textbf{Pound} \\
\midrule
\textbf{Size} & Small (S) & Medium (M) & Large (L) & Extra Large (XL) \\
\textbf{Industry} & Tech & Media & Pharmacare & Finance \\
\midrule
\textbf{Entity Types} \\
Task & 103 & 1,134 & 6,479 & 52,504 \\
Meeting & 76 & 890 & 6,445 &  57,265\\
Employee & 12 & 122 & 1,110 & 10,210 \\
Project & 7 & 77 & 317 & 587 \\
Team & 1 & 11 & 90 & 207 \\
Department & 1 & 7 & 11 & 15 \\
Company & 1 & 1 & 1 & 1 \\
\midrule
\textbf{Relationship Types} \\
WorksOn & 124 & 1,356 & 7,708 &  63,431\\
ReportsTo & 11 & 133 & 1,255 & 11,455 \\
MemberOf & 11 & 126 & 1,576 & 15,954 \\
WorksAt & 12 & 122 & 1,110 & 10,210 \\
Attends & 221 & 2,709 & 22,772 & 213,590 \\
Organize & 76 & 890 & 6,445 & 57,265 \\
\midrule
\textbf{Total Entities} & 201 & 2,242 & 14,453 & 120,789 \\
\textbf{Total Relationships} & 455 & 5,336 & 40,866 & 371,905 \\
\textbf{Total Documents} & 354 & 3,926 & 26,493 & 232,693 \\
\bottomrule
\end{tabular*}
\caption{Summary statistics for each of the four synthetic companies generated for \textsc{CorporateBench}. Note that each company represents an additional order of magnitude over the previous in all respects.}
\label{tab:company_datasets}
\end{table*}
\renewcommand{\arraystretch}{1.0}

We use the pipeline described in \autoref{sec:construction} to produce four company knowledge bases with $10^1$ to  $10^4$ employees.
We present these companies in \autoref{tab:company_datasets}.
We validate the dataset in four ways:

    \paragraph{\textbf{Network properties.}}
    We observe consistent scaling patterns across each company size.
    For example, the smallest company (Zenith Labs) contains 12 employees, while the largest company (Pound) contains 10,210 individuals.     Zenith Labs contains 455 total relationships at an average $\approx$38 degrees per employee; while Pound contains $\approx$37 degrees.
    This stable scaling behaviour helps downstream tasks control for issues arising from poor scaling.
    \paragraph{\textbf{Examining sampled documents.}}
    We generate a total 263,466 documents across all four company KBs to produce four corpora containing, respectively: 354 documents, 3,926 documents, 26,493 documents, and 232,693 documents.
    Each document explicitly reveals one or more relationships in each company KB following the sampling methodology described in \autoref{subsec:documents}.\footnote{We provide representative sample documents for each QA type in Appendices~G--I, illustrating document diversity, question complexity, and ground-truth answer precision.}
    \paragraph{\textbf{Validating document topics.}}
    We provide topic distributions for each company corpus in \autoref{tab:company_topics}.
    Total topic counts scale linearly from 6 in Zenith (S), 60 for Streamvibe (M), 600 for Biocure (L), and finally 6,000 topics in Pound (XL).
    \paragraph{\textbf{Human validation.}}
    We sample 100 random emails from the full \textsc{CorporateBench} corpus for human validation.
    Because each document positively asserts a relationship (e.g. that one person manages another), we prompt Claude Sonnet 4 to invert each document by modifying the evidence string to negate the relationship.
    This yields a balanced test set of positive and negative examples.
    We pair ten groups of five non-expert reviewers with batches of 20 documents each, asking them to judge whether each document positively asserts the relationship on a binary scale.
    Across all 1,000 judgements, reviewers correctly identify the intended relationship 76.2\% of the time.

%% file: writing/benchmark.tex
\begin{table}[t]
    \centering\
    \small
    \begin{tabular}{m{0.20\columnwidth} m{0.60\columnwidth}}
        \textbf{KB QA} & How many people began working at Pound Financial Group LLC after 7th March 2024? \\
        \midrule
        \textbf{Topic QA} & Did Marion Friend discuss both transportation and value creation in communications after 30 Jan 2024? \\
        \midrule
        \textbf{Integrated QA} & Who among client financial documentation preparation workers sent external emails about due diligence? \\
    \end{tabular}
    \caption{Example questions for Pound (XL) across KB, Topic, and Integrated QA. More extensive question breakdowns are available in \autoref{app:qa-benchmarks}.}
    \label{tab:qa-example-questions}
\end{table}

\textsc{CorporateBench} tests models on five tasks organized in two categories: extraction and QA. The two assess distinct capabilities. Extraction assesses whether a model can \textit{construct} a structured knowledge base from raw documents while QA evaluates \textit{reasoning} over an already-constructed representation. A model may answer KB QA questions correctly yet fail to build the KB in the first place. 

\subsection{Extraction Tasks}
Our two extraction tasks evaluate models on their ability to recover (i) relations and (ii) topics from each company corpus. The tasks are:
\paragraph{KB Evaluation.}
This task involves three stages: ingestion, mapping, and evaluation. 
During ingestion, LLMs extract entities and relationships from documents using the prompt template in \autoref{app:extraction-prompt}. 
We then apply deduplication techniques like exact name match and email-based clustering to reduce redundancy.
This process yields triples in the form $\langle$subject, predicate, object$\rangle$.
We evaluate these ingested triples against ground truth triples from \textsc{CorporateBench} and compute $F_1$ for all matched entities/relationships.

\paragraph{Topic Classification.}
This task tests the model's ability to classify email topics via multi-class classification on a subset of topics from the dataset.
The model must select one of either 31 classes (for the large company \textit{Biocure}) or 17 classes (for the extra large company \textit{Pound}) on a stratified test set of 1000 documents per corpus as provided in \autoref{app:topics}.
We subset these topics from the original full set of topics by navigating up the topic hierarchy as described in \autoref{app:topic-levels} to arrive at a lower count tenable for classification with current LLMs.\footnote{We originally defined 31 topics for Pound as well, reducing it to 17 topics for evaluation due to topic overlap.}
We finally evaluate performance with a macro $F_1$ averaged across the classes.

\begin{figure*}[t]
    \centering
    \includegraphics[width=1\linewidth]{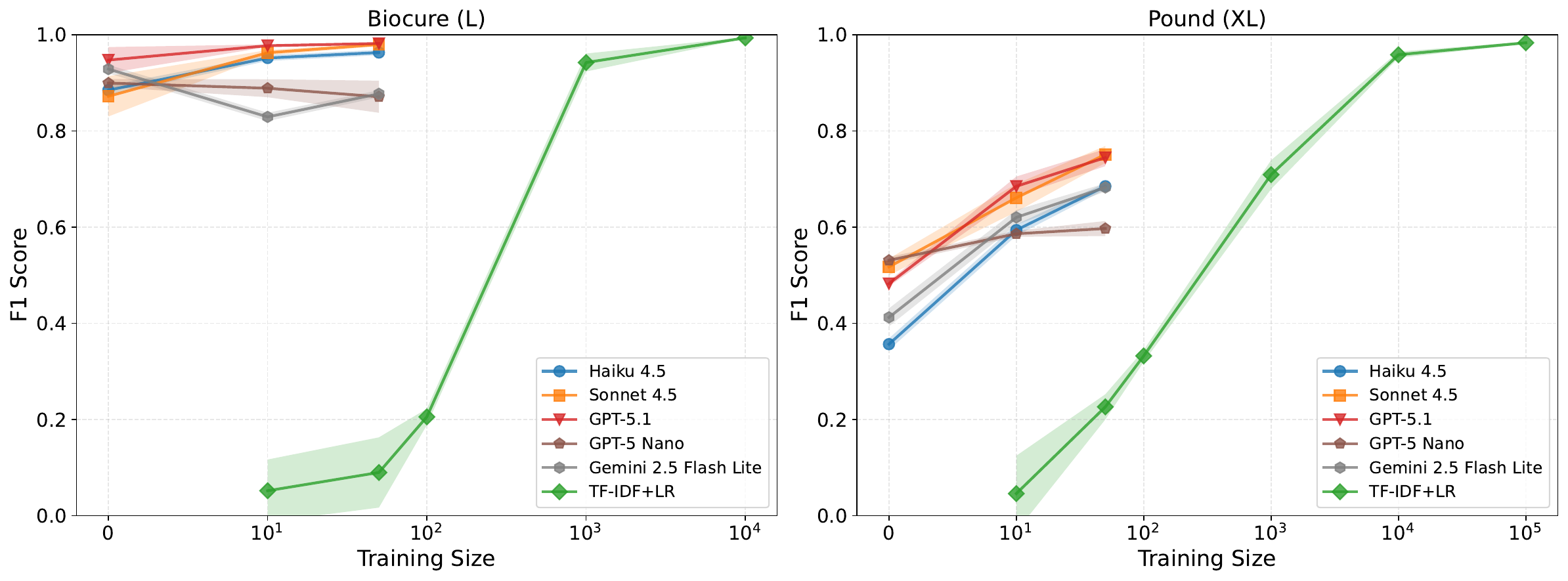}
    \caption{Topic classification $F_1$ scores across number of few-shot examples provided in the prompt (for LLMs) or labeled training examples (for TF-IDF+LR) for Biocure (L) and Pound (XL) datasets.}
    \label{fig:topic-classification}
\end{figure*}

\begin{table*}[t]
\centering
\footnotesize
\begin{tabular}{llccccccccc}
\toprule
& & \multicolumn{3}{c}{\textbf{Entities}}
& \multicolumn{3}{c}{\textbf{Relationships}}
& \multicolumn{3}{c}{\textbf{Temporal Relationships}$^{\dagger}$} \\
\cmidrule(lr){3-5} \cmidrule(lr){6-8} \cmidrule(lr){9-11}
\textbf{Size} & \textbf{Model} & \textbf{Precision} & \textbf{Recall} & \textbf{F$_1$} & \textbf{Precision} & \textbf{Recall} & \textbf{F$_1$} & \textbf{Precision} & \textbf{Recall} & \textbf{F$_1$} \\
\midrule
\multirow{3}{*}{S} & Haiku 4.5 & .866 & .786 & \textbf{.824} & .751 & .966 & \textbf{.845} & .453 & .488 & \textbf{.470} \\
& GPT-5 Nano & .926 & .624 & .746 & .795 & .285 & .419 & .317 & .092 & .142 \\
& Gemini 2.5 F-L & .846 & .792 & .818 & .677 & .969 & .797 & .441 & .444 & .442 \\
\midrule
\multirow{3}{*}{M} & Haiku 4.5 & .911 & .721 & \textbf{.805} & .782 & .595 & \textbf{.676} & .427 & .209 & .281 \\
& GPT-5 Nano & .902 & .645 & .752 & .569 & .165 & .256 & .273 & .046 & .078 \\
& Gemini 2.5 F-L & .849 & .749 & .796 & .613 & .708 & .657 & .391 & .279 & \textbf{.326} \\
\midrule
\multirow{3}{*}{L} & Haiku 4.5 & .902 & .704 & \textbf{.790} & .646 & .556 & \textbf{.597} & .390 & .207 & \textbf{.271} \\
& GPT-5 Nano & .928 & .673 & .780 & .479 & .167 & .247 & .254 & .055 & .090 \\
& Gemini 2.5 F-L & .825 & .707 & .762 & .417 & .539 & .470 & .010 & .012 & .011 \\
\midrule
\multirow{3}{*}{XL} & Haiku 4.5 & .937 & .648 & .766 & .616 & .412 & \textbf{.494} & .310 & .120 & \textbf{.173} \\
& GPT-5 Nano & .962 & .652 & \textbf{.777} & .456 & .133 & .206 & .191 & .037 & .062 \\
& Gemini 2.5 F-L & .899 & .594 & .715 & .227 & .186 & .205 & .289 & .082 & .128 \\
\bottomrule
\end{tabular}
\caption{KB Evaluation Results. \textbf{Bold} denotes best score out of the three models.}
\label{tab:kb-eval-results}
\end{table*}

\begin{table*}[t]
\centering
\footnotesize
\begin{tabular}{@{}llcccccccccccc@{}}
\toprule
\multirow{2}{*}{\textbf{Method}} & \multirow{2}{*}{\textbf{Model}} & \multicolumn{4}{c}{\textbf{KB QA}} & \multicolumn{4}{c}{\textbf{Topic QA}} & \multicolumn{4}{c}{\textbf{Integrated QA}} \\
\cmidrule(lr){3-6} \cmidrule(lr){7-10} \cmidrule(lr){11-14}
  &   & \textbf{S} & \textbf{M} & \textbf{L} & \textbf{XL} & \textbf{S} & \textbf{M} & \textbf{L} & \textbf{XL} & \textbf{S} & \textbf{M} & \textbf{L} & \textbf{XL} \\
\midrule
\multirow{5}{*}{RAG} & Haiku 4.5 & .265 & .273 & .189 & .164 & .650 & \textbf{.563} & .418 & .464 & .332 & .277 & .327 & .424 \\
  & Sonnet 4.5 & .322 & .311 & .198 & .215 & .656 & .535 & \textbf{.423} & \textbf{.492} & .439 & .250 & .272 & .401 \\
  & GPT-5 Nano & \textbf{.527} & \textbf{.501} & \textbf{.312} & .244 & \textbf{.672} & .479 & .385 & .456 & \textbf{.511} & .475 & .468 & .510 \\
  & GPT-5.1 & .470 & .440 & .276 & \textbf{.269} & .634 & .485 & .406 & .456 & .469 & \textbf{.564} & \textbf{.560} & \textbf{.566} \\
  & Gemini 2.5 F-L & .230 & .200 & .199 & .147 & .572 & .401 & .374 & .376 & .338 & .218 & .313 & .295 \\
\midrule
\multirow{5}{*}{KB} & Haiku 4.5 & .702 & .724 & .614 & .597 & .811 & .619 & .517 & .488 & .488 & .540 & .538 & .540 \\
  & Sonnet 4.5 & \textbf{.767} & \textbf{.742} & \textbf{.628} & \textbf{.642} & \textbf{.848} & \textbf{.687} & \textbf{.576} & \textbf{.571} & \textbf{.611} & .624 & .644 & \textbf{.647} \\
  & GPT-5 Nano & .577 & .484 & .519 & .522 & .528 & .431 & .384 & .409 & .434 & .611 & .588 & .567 \\
  & GPT-5.1 & .555 & .547 & .481 & .525 & .468 & .403 & .387 & .397 & .431 & \textbf{.640} & \textbf{.653} & .607 \\
  & Gemini 2.5 F-L & .258 & .207 & .248 & .192 & .384 & .384 & .304 & .352 & .272 & .483 & .528 & .412 \\
\bottomrule
\end{tabular}
\caption{KB QA, Topic QA, and Integrated QA Results. \textbf{Bold} denotes best model performance for the particular QA task using the respective methods (RAG or KB).}
\label{tab:qa-results}
\end{table*}

\subsection{QA Tasks}
Our three QA tasks evaluate models on factual questions about the company and its documents.
QA questions are generated from 250 manually written question templates instantiated across companies and tasks with placeholder variables (e.g., employee or task) that are filled dynamically with values from the KB via verified SPARQL queries (and thus not by LLM generation).
By construction, more complex questions require gathering information across more nodes in the knowledge graph and thus across more documents.
QA questions come in five answer types: string, date, integer, boolean, and sets of strings.
All types, except for sets, are judged on exact match (1 if match, 0 if not), but we calculate $F_1$ for sets to allow models partial credit.
We compare models in two QA settings. 
The first is \textbf{RAG} \citep{lewis2021retrievalaugmentedgenerationknowledgeintensivenlp}. 
We equip models with a RAG-based search tool containing all the documents.\footnote{Using \texttt{pgvector} and \texttt{text-embedding-3-small}.} 
This setting establishes performance in realistic corporate settings. 
The next setting is \textbf{KB}, where we equip models with a SQL tool for querying data directly imported with our ETL pipeline from our ground truth KB file.\footnote{We describe the pipeline in \autoref{app:etl-process}.}
This setting establishes an upper bound on model performance \textit{relative to the RAG setting}, since models receive direct access to structured ground truth rather than retrieving from noisy documents.
We note that text-to-SQL remains a challenging task in its own right, and that these scores do not represent an absolute performance ceiling.
The tasks are as follows:

\paragraph{KB QA.}
This task involves answering questions about entities, relationships, and the temporality of these relationships.
The example question in \autoref{tab:qa-example-questions} requires identifying employees from emails, \textit{and} temporal reasoning to identify when employees joined based on onboarding documents.

\paragraph{Topic QA.}
This task involves answering questions about the topics of the documents that are covered in emails. 
The Topic QA example in \autoref{tab:qa-example-questions} requires models to read the employee's emails during a specific interval of time and answer whether the employee discussed that topic.

\paragraph{Integrated QA.}
The final QA task combines KB QA and Topic QA.
The example question in \autoref{tab:qa-example-questions} requires models to determine which employees have worked on a task and \textit{then} analyze the content of specific emails they sent.

%% file: writing/results_new.tex
We evaluate five contemporary models on \textsc{CorporateBench} to provide a sense of the current state of the art.
These models are: Claude Haiku 4.5, Claude Sonnet 4.5, GPT-5 Nano, GPT-5.1, and Gemini 2.5 Flash Lite.
We specifically select ``lighter'' models to strike a balance between performance and cost given evaluating a model on the full 263,466 document set in \textsc{CB} can incur significant API costs.
We provide a full run-down of our evaluation metrics in \autoref{app:experimental_setup}.

\subsection{Extraction Tasks}

For KB evaluation, Table~\ref{tab:kb-eval-results} shows entity extraction substantially outperforms relationship extraction across all models and dataset sizes. Entity extraction remains stable across scales ($F_1$: 0.715–0.824), while relationship extraction degrades notably from smaller (S: 0.419–0.845, M: 0.256–0.676) to larger datasets (L: 0.247–0.597, XL: 0.205–0.494). Temporal relationships prove most challenging, deteriorating from $F_1$ 0.142–0.470 (S) to 0.062–0.173 (XL).
These findings indicate models struggle to maintain relational consistency at scale, even as factual consistency (entity extraction) is preserved.
Error analysis revealed two primary causes: (1) entity extraction errors cascade into relationship extraction, and (2) contradictory relationship descriptions accumulate across documents, with each model struggling with different relationship types.\footnote{See \autoref{app:kb-eval-results} for more information.}

For topic classification, we compare against a TF-IDF+LR baseline.
Figure~\ref{fig:topic-classification} and Table~\ref{tab:topic-classification-results} reveal distinct patterns across datasets.
On Biocure (L), LLMs achieve strong few-shot performance ($F_1$: 0.872--0.950 zero-shot, 0.877--0.982 with 50 examples), matching TF-IDF+LR's 10K-example performance ($F_1$: 0.993) with 200$\times$ less data.
On Pound (XL), LLMs show steeper learning curves ($F_1$: 0.357--0.531 zero-shot to 0.598--0.751 with 50 examples) but underperform the data-rich baseline ($F_1$: 0.983 at 100K examples).
LLMs performed better on Biocure (31 categories) than Pound (17 categories) despite more classes, likely because Pound's hierarchical topic merging created ambiguous boundaries and semantic overlap.

\subsection{QA Tasks}

We report all QA performance in \autoref{tab:qa-results}. Models consistently perform better with direct KB access than with RAG.
Our design holds the number of questions constant at 250 per company per task (\autoref{app:experimental_setup}) while increasing corpus size, so larger companies present proportionally more documents per question ($\approx$1.4 to $\approx$931).
Across all three QA types, the best KB scores exceed the corresponding best RAG scores, and this gap widens with scale: on KB QA, the KB--RAG difference increases from 0.24 (S) to 0.37 (XL).
This suggests RAG may suffice for smaller corpora, but scale causes problems when surfacing knowledge.

Performance degrades with corpus size for KB QA and Topic QA, reflecting the difficulty of locating precise information in larger collections.
Integrated QA exhibits a different pattern: performance slightly improves with scale for both methods, likely because larger corpora provide richer cross-source context for synthesis.

We finally note model-level differences: both GPT-5 models outperform others with RAG, while Claude Sonnet 4.5 excels at KB querying.
However, GPT-5.1 exhibits an early stopping problem that disproportionately affects the KB setting (191 occurrences vs.\ 56 with RAG), partially explaining the performance gap between methods.\footnote{Specifically, the model returns empty string after receiving a tool response instead of generating an answer.}

%% file: writing/conclusion.tex
We release \textsc{CorporateBench}, a benchmark for testing LLMs on enterprise-specific retrieval and reasoning tasks.
\textsc{CB} features 263,466 total synthetic documents sampled from four procedurally generated knowledge bases representing companies of varying scale.
This novel generation process allows \textsc{CB} to produce realistic tasks over corpora of arbitrary scale while maintaining inter-document logical consistency, a quality predecessor enterprise benchmarks lacked.
Testing five recent LLMs from Anthropic, OpenAI, and Google indicates contemporary LLMs are marginally competent at completing enterprise tasks, with performance decreasing inversely with corpora scale up to and beyond $10^5$ documents.
Our methods, benchmark, and results are of value to researchers and industry practitioners interested in assessing LLM performance when deployed in corporate environments.
\paragraph{Next steps.}
We encourage future researchers to continue closing the gap between synthetic and realistic benchmarks.
Researchers working in other domains (e.g. deep research) will want to evaluate whether replicating our use of procedurally generated KBs will be similarly fruitful.

%% file: writing/ethics.tex
All synthetically generated materials in \textsc{CorporateBench} were generated via OpenAI and Anthropic-provided API endpoints with content filtering enabled.
We note LLM developers often use benchmarks as a reference point during training, and that this practice can lead to benchmark-specific idiosyncrasies and can disproportionately emphasize certain semantic and/or syntactic qualities in model output.
We therefore promoted ethnic and racial diversity in our generated companies by pre-computing names with \href{https://pypi.org/project/Faker/}{Faker}, a Python library for procedurally generating personal information like names and emails.
We enable all Latin alphabet-based Faker lists, ensuring a fair and diverse sample of humans across all four companies.

%% file: writing/appendix.tex
\clearpage

\section{Company Information Overview}
\label{app:worlds}

\autoref{tab:company_data_1} contains information about the four companies that we have generated. We have generated four companies with different attributes: name, industry, size, model, and story.

\begin{table*}[htbp]
\centering
\footnotesize
\begin{tabular}{llllp{9.5cm}}
\toprule
\textbf{Name} & \textbf{Industry} & \textbf{Size} & \textbf{Model} & \textbf{Story} \\
\midrule
Zenith & Tech & Small & B2C & Nestled in the heart of California's tech scene, Zenith Labs is a nimble, people-first B2C company crafting sleek laptops, phones, and desktops for everyday users who crave simplicity without sacrificing style. With few employees and a laid-back, startup vibe, Zenith Labs punches above its weight in the consumer electronics space, offering intuitive, affordable devices that blend minimalist design with reliable performance. Their audience? Young professionals, students, and creatives who want tech that feels personal—not corporate. Zenith Labs' motto "Keep it crisp" reflects their mission to strip away the noise and deliver clean, user-friendly experiences. Their goal is to become the go-to brand for people who want smart tech that doesn't try too hard. \\
\midrule
Streamvibe & Media & Medium & B2C & A creative powerhouse in the entertainment industry, StreamVibe Studios produces original streaming content, podcasts, and digital campaigns for the cord-cutting generation. With talented creators, producers, and marketers spread across Los Angeles and Austin, StreamVibe has carved out a niche creating viral web series, documentary films, and branded content for Gen Z and millennial audiences. Known for their "content first, suits second" mentality, StreamVibe attracts top creative talent with flexible work arrangements, profit-sharing, and a culture that celebrates bold storytelling and authentic voices. \\
\midrule
Biocure & Pharma & Large & B2B & Operating from a sprawling research campus with state-of-the-art laboratories, BioCure Solutions is a science-driven B2B pharmaceutical company specializing in contract research and development services for drug discovery and clinical trials. This established industry leader employs scientists, researchers, and regulatory specialists who maintain rigorous FDA compliance standards and operate under strict Good Manufacturing Practice protocols that have secured partnerships with major pharmaceutical corporations worldwide. Their mission is to become the premier CRO partner for pharmaceutical companies developing tomorrow's therapies, providing the scientific rigor and regulatory expertise that transforms promising compounds into approved medicines that improve human health globally.\\
\midrule
Pound & Finance & X-Large & B2B & Pound Financial is a mid-market investment firm specializing in private equity and corporate restructuring for manufacturing and technology companies. The firm maintains a conservative, relationship-driven approach, focusing on long-term value creation rather than quick exits. Their clientele includes family-owned businesses seeking growth capital and Fortune 500 companies navigating complex financial transitions. Pound Financial's reputation for discretion, thorough due diligence, and strategic guidance has made them a trusted partner for executives facing critical financial decisions. \\
\bottomrule
\end{tabular}
\caption{Company/KB Information Overview.}
\label{tab:company_data_1}
\end{table*}

\section{Evidence Strings}
\label{app:evidence-strings}

\paragraph{Non-meeting evidence strings.} These strings substantiate relationships between email senders and entities (teams, managers, tasks, or companies) through linguistic patterns that vary in explicitness and temporal stage. 

Each relationship type contains evidence strings organized by temporal stage (start, ongoing, end) and evidence strength (explicit, implicit).
Explicit strings directly state the relationship using clear relationship verbs, while implicit strings demonstrate active engagement through actions, coordination, or possessive language.

To ensure lexical diversity within implicit strings, we organize them into thematic subcategories that represent different aspects of workplace behavior.

For example, the \texttt{memberOf} relationship includes implicit subcategories such as "onboarding\_to\_team" for new members learning processes, "syncing\_with\_team\_members" for ongoing coordination, and "knowledge\_transfer\_to\_team" for members departing the team.
The \texttt{memberOf} relationship contains 20 explicit strings per temporal stage and 140 implicit strings (7 subcategories with 20 strings each) per temporal stage, totaling 480 strings.
Similar distributions apply to other relationship types, resulting in 3,217 total evidence string variations across all six relationship types.

\autoref{fig:member-of-evidence-strings} shows example evidence strings for the \texttt{memberOf} relationship.

\paragraph{Meeting evidence strings.} These evidence strings differ from non-meeting strings by focusing on task and project progress rather than relationship formation or dissolution.

These strings are organized by progress level rather than temporal stage, with four progress bins: 0-25\%, 25-50\%, 50-75\%, and 75-100\%.
Each progress bin contains both explicit strings that directly state completion percentage and implicit strings organized into thematic subcategories representing typical activities at that stage of work.

For example, the 0-25\% progress bin includes implicit subcategories such as "initial\_research" for gathering requirements, "project\_kickoff" for launching initiatives, and "resource\_gathering" for assembling team and materials.
As progress advances, the subcategories shift to reflect later-stage activities: the 50-75\% bin includes "beta\_testing" and "integration\_work", while the 75-100\% bin includes "final\_review", "deployment\_prep", and "closing\_activities".

Each progress bin contains 20 explicit strings and 120 implicit strings (6 subcategories with 20 strings each), totaling 560 strings for the \texttt{taskProgress} evidence type.

\autoref{fig:task-progress-evidence-strings} shows example evidence strings for different progress levels.

\begin{table*}[t]
\small
\centering
\begin{tabular}{@{}lll@{}}
\toprule
\textbf{Temporal Stage} & \textbf{Strength} & \textbf{Examples} \\
\midrule
\multirow{8}{*}{Start} 
  & \multirow{4}{*}{Explicit} 
    & I just joined \{\{object\}\} and wanted to introduce myself \\
  & & I'm excited to announce that I'm now part of \{\{object\}\} \\
  & & I recently became a member of \{\{object\}\} \\
  & & \emph{... 17 more} \\
\cmidrule{2-3}
  & \multirow{4}{*}{Implicit} 
    & \textbf{onboarding\_to\_team:} Just going through the \{\{object\}\} onboarding materials today \\
  & & \textbf{learning\_team\_processes:} Still getting familiar with how \{\{object\}\} runs things \\
  & & \textbf{getting\_added\_to\_distribution\_lists:} Just got added to the \{\{object\}\} email lists \\
  & & \emph{... 4 more subcategories, 20 strings each} \\
\midrule
\multirow{8}{*}{Ongoing} 
  & \multirow{4}{*}{Explicit} 
    & As a member of \{\{object\}\}, I wanted to share our latest findings \\
  & & I'm on \{\{object\}\}, and we've been tracking this issue \\
  & & My group, \{\{object\}\}, has identified a solution \\
  & & \emph{... 17 more} \\
\cmidrule{2-3}
  & \multirow{4}{*}{Implicit} 
    & \textbf{discussing\_team\_priorities:} This aligns with \{\{object\}\}'s Q1 priorities \\
  & & \textbf{syncing\_with\_team\_members:} Let me check with my \{\{object\}\} teammates first \\
  & & \textbf{contributing\_to\_team\_decisions:} \{\{object\}\} voted on this approach last week \\
  & & \emph{... 4 more subcategories, 20 strings each} \\
\midrule
\multirow{8}{*}{End} 
  & \multirow{4}{*}{Explicit} 
    & I'm leaving \{\{object\}\}, effective today \\
  & & My time with \{\{object\}\} is coming to an end \\
  & & I'm departing from \{\{object\}\} today \\
  & & \emph{... 17 more} \\
\cmidrule{2-3}
  & \multirow{4}{*}{Implicit} 
    & \textbf{knowledge\_transfer\_to\_team:} Documenting everything for \{\{object\}\} before I transition off \\
  & & \textbf{handing\_off\_team\_responsibilities:} Transitioning my \{\{object\}\} responsibilities to Sarah \\
  & & \textbf{wrapping\_up\_team\_projects:} Finishing up my last \{\{object\}\} project this week \\
  & & \emph{... 4 more subcategories, 20 strings each} \\
\bottomrule
\end{tabular}
\caption{Example evidence strings for the \texttt{memberOf} relationship showing temporal stages and evidence strength levels. Explicit strings directly state team membership, while implicit strings demonstrate engagement through workplace actions. The placeholder \texttt{\{\{object\}\}} represents the team name.}
\label{fig:member-of-evidence-strings}
\end{table*}

\begin{table*}[t]
\small
\centering
\begin{tabular}{@{}lll@{}}
\toprule
\textbf{Progress Level} & \textbf{Strength} & \textbf{Examples} \\
\midrule
\multirow{8}{*}{0-25\%} 
  & \multirow{4}{*}{Explicit} 
    & \{\{employee\}\} \{\{have\_verb\}\} \{\{task\}\} about 20\% done \\
  & & \{\{employee\}\} \{\{be\_verb\}\} roughly a quarter of the way through \{\{task\}\} \\
  & & \{\{task\}\} is still in early stages with \{\{employee\}\}, around 15-25\% complete \\
  & & \emph{... 17 more} \\
\cmidrule{2-3}
  & \multirow{4}{*}{Implicit} 
    & \textbf{initial\_research:} \{\{employee\}\} \{\{be\_verb\}\} still gathering requirements for \{\{task\}\} \\
  & & \textbf{project\_kickoff:} \{\{employee\}\} sent out the project charter for \{\{task\}\} today \\
  & & \textbf{early\_planning:} \{\{employee\}\} \{\{be\_verb\}\} sketching out the roadmap for \{\{task\}\} \\
  & & \emph{... 3 more subcategories, 20 strings each} \\
\midrule
\multirow{8}{*}{25-50\%} 
  & \multirow{4}{*}{Explicit} 
    & \{\{employee\}\} \{\{have\_verb\}\} \{\{task\}\} about 40\% complete \\
  & & \{\{employee\}\} \{\{be\_verb\}\} roughly a third done with \{\{task\}\} \\
  & & \{\{task\}\} is coming along with \{\{employee\}\}, around 35\% finished \\
  & & \emph{... 17 more} \\
\cmidrule{2-3}
  & \multirow{4}{*}{Implicit} 
    & \textbf{prototype\_development:} \{\{employee\}\} just got the first prototype working for \{\{task\}\} \\
  & & \textbf{foundation\_complete:} \{\{employee\}\} \{\{have\_verb\}\} the core framework built for \{\{task\}\} \\
  & & \textbf{initial\_deliverables:} \{\{employee\}\} submitted the first milestone deliverable for \{\{task\}\} \\
  & & \emph{... 3 more subcategories, 20 strings each} \\
\midrule
\multirow{8}{*}{50-75\%} 
  & \multirow{4}{*}{Explicit} 
    & \{\{employee\}\} \{\{have\_verb\}\} \{\{task\}\} about 60\% done now \\
  & & \{\{employee\}\} \{\{be\_verb\}\} past halfway on \{\{task\}\}, around 65\% complete \\
  & & \{\{task\}\} is roughly two-thirds finished by \{\{employee\}\} \\
  & & \emph{... 17 more} \\
\cmidrule{2-3}
  & \multirow{4}{*}{Implicit} 
    & \textbf{beta\_testing:} \{\{employee\}\} \{\{be\_verb\}\} running trial programs for \{\{task\}\} \\
  & & \textbf{debugging\_phase:} \{\{employee\}\} \{\{be\_verb\}\} fixing the remaining problems in \{\{task\}\} \\
  & & \textbf{integration\_work:} \{\{employee\}\} \{\{be\_verb\}\} bringing together all the pieces of \{\{task\}\} \\
  & & \emph{... 3 more subcategories, 20 strings each} \\
\midrule
\multirow{8}{*}{75-100\%} 
  & \multirow{4}{*}{Explicit} 
    & \{\{employee\}\} \{\{have\_verb\}\} \{\{task\}\} nearly done, about 85\% complete \\
  & & \{\{employee\}\} \{\{be\_verb\}\} almost finished with \{\{task\}\}, around 80-90\% there \\
  & & \{\{task\}\} is in the final stretch with \{\{employee\}\}, roughly 80\% done \\
  & & \emph{... 17 more} \\
\cmidrule{2-3}
  & \multirow{4}{*}{Implicit} 
    & \textbf{final\_review:} \{\{employee\}\} \{\{be\_verb\}\} doing the final review of \{\{task\}\} \\
  & & \textbf{polishing\_details:} \{\{employee\}\} \{\{be\_verb\}\} polishing the documentation for \{\{task\}\} \\
  & & \textbf{deployment\_prep:} \{\{employee\}\} \{\{be\_verb\}\} preparing \{\{task\}\} for implementation \\
  & & \emph{... 3 more subcategories, 20 strings each} \\
\bottomrule
\end{tabular}
\caption{Example evidence strings for the \texttt{taskProgress} relationship showing progress levels and evidence strength. Explicit strings directly state completion percentages, while implicit strings demonstrate typical activities at each stage. The placeholders \texttt{\{\{employee\}\}} and \texttt{\{\{task\}\}} represent the employee name and task name respectively.}
\label{fig:task-progress-evidence-strings}
\end{table*}

\section{Topics}
\label{app:topic-levels}

\renewcommand{\arraystretch}{0.7}
\begin{table}[htbp]
\centering
\footnotesize
\begin{tabular}{cccc}
\toprule
\textbf{Size} & \textbf{Documents} & \textbf{Non-Business} & \textbf{Business} \\
\midrule
\textbf{S} & $\approx 4 \cdot 10^2$ & $3 \cdot 10^0$ & $3 \cdot 10^0$\\
\textbf{M} & $\approx 4 \cdot 10^3$ & $3 \cdot 10^1$ & $3 \cdot 10^1$ \\
\textbf{L} & $\approx 3 \cdot 10^4$ & $3 \cdot 10^2$ & $3 \cdot 10^2$ \\
\textbf{XL} & $\approx 2 \cdot 10^5$ & $3 \cdot 10^3$ & $3 \cdot 10^3$ \\
\bottomrule
\end{tabular}
\caption{Breakdown of the number of topics used to condition the generation of documents in \textsc{CorporateBench} based on different sizes of the company.  We provide more detail in \autoref{app:topic-levels}.}
\label{tab:company_topics}
\end{table}
\renewcommand{\arraystretch}{1.0}

\paragraph{Non-meeting topics.} 
For non-meeting documents, we assign topics related to either business or non-business operations. 
To maintain lexical diversity as the corpus grows, we scale the number of topics proportionally by creating hierarchical topic structures of varying depth, similar to the scenario-seeded approach defined in \citep{sun-etal-2024-conscendi}.

We begin with broad top-level categories. 
Non-business topics consistently use three main categories: "Food", "Travel", and "Transportation". 
Business topics vary by company domain: for example, Streamvibe (media) uses "Content Creation", "Production and Distribution", and "Audience Engagement", while Biocure (pharmaceuticals) uses "Drug Development", "Drug Approval", and "Drug Sales", and Pound (finance) uses "Deal Sourcing", "Due Diligence", and "Value Creation".

The hierarchy depth depends on corpus size. 
Zenith, our smallest company, requires no subcategories—its business topics are specific from the start (e.g., "API Security Standards", "Microservice Architecture Patterns"), as are its non-business topics (e.g., "Home Cooking", "Vacation Plans"). 
For larger companies like Streamvibe, Biocure, and Pound, we recursively expand top-level categories into deeper hierarchies using Claude Sonnet 4.5, creating subcategories at two, three, or four levels as needed. 
We manually review each generated topic to ensure it is appropriate for corporate email content before inclusion. 

\autoref{fig:biocure-topics} shows the hierarchical example of topics for Biocure, our large company. 

\begin{figure*}
    \centering
    \includegraphics[width=\linewidth]{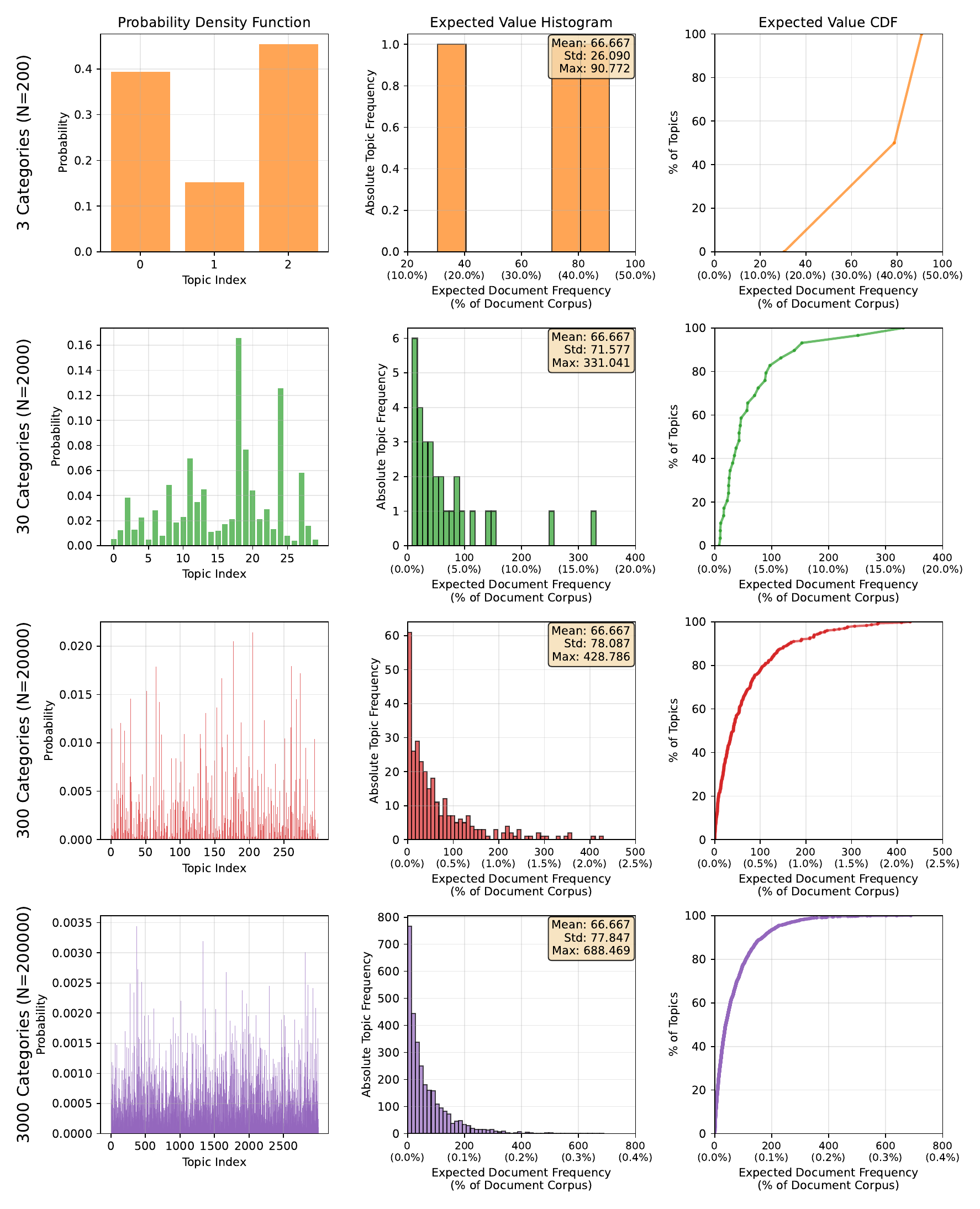}
    \caption{Summary of theoretical Dirichlet distributions used for topic sampling when generating non-meeting documents.}
    \label{fig:dirichlet_topic_distributions}
\end{figure*}

\begin{figure*}
    \centering
    \includegraphics[width=\linewidth]{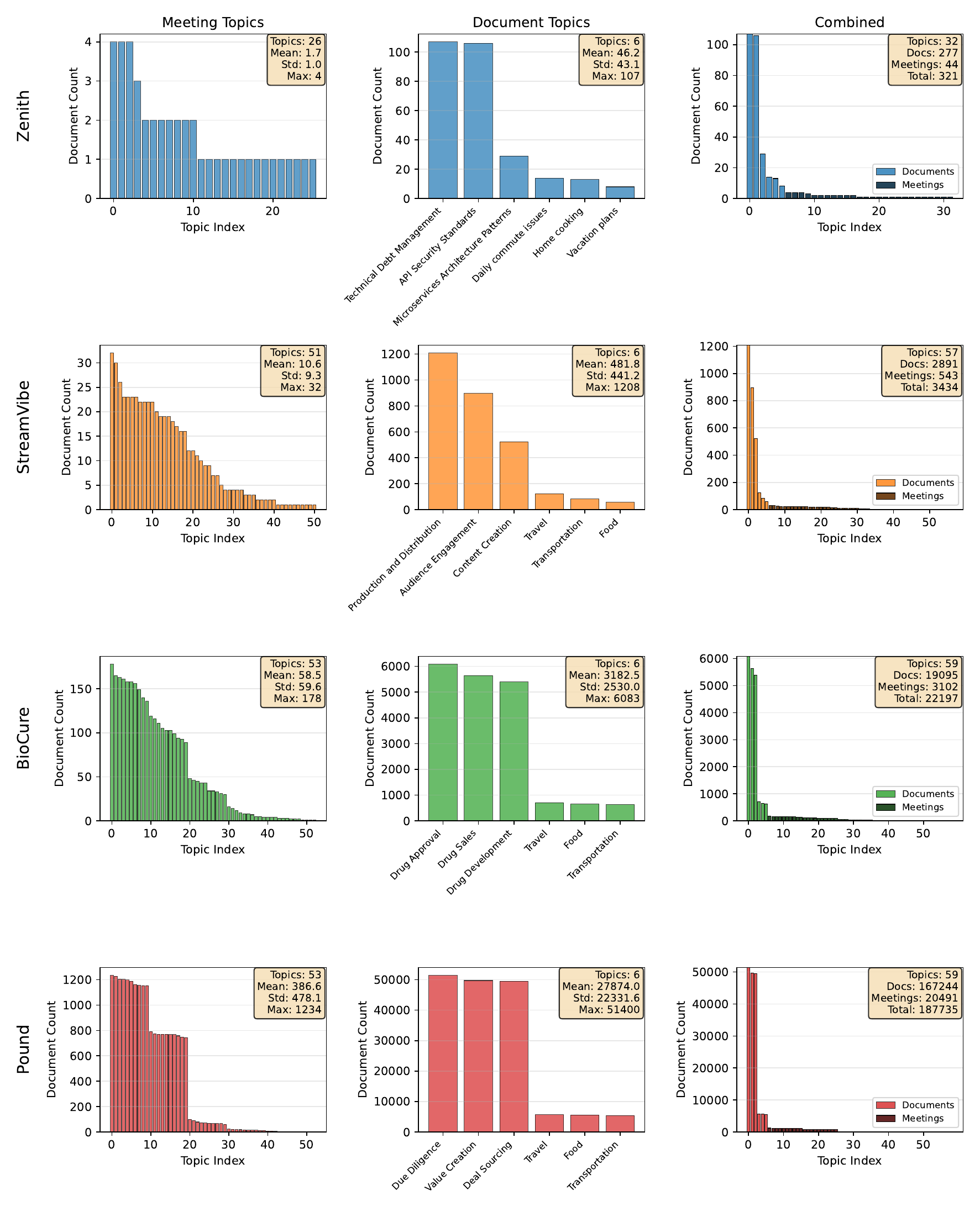}
    \caption{Summary of empirical topic distributions of generated documents across all four KBs. Includes meeting and non-meeting topics. See \autoref{app:topic-levels} for more details on topics.}
    \label{fig:actual_topic_distributions}
\end{figure*}

Finally, to select non-meeting topics, we use a Dirichlet topic distribution with a sparse alpha concentration of 0.75 to determine the probability at which we'd use a topic for a particular document. 
These distributions are illustrated in \autoref{fig:dirichlet_topic_distributions}. 
The actual topic distributions after generation are illustrated in \autoref{fig:actual_topic_distributions}.

\paragraph{Meeting topics.}
Meeting documents require a different approach to topic assignment, as their content is shaped by meeting dynamics rather than general business operations. 
We select meeting topics from a separate pool, conditioned on meeting type to reflect typical discussion patterns. 
For example, direct report meetings commonly cover "Employee Performance Review", while executive meetings focus on "Company Strategy and Vision Alignment".

Not all meeting emails receive topic assignments. 
Blank calendar invites and similar minimal-content emails lack substantive discussion and are therefore excluded from topic assignment and from our Topic Classification, Topic QA, and Integrated QA evaluation tasks.

\autoref{table:meeting-topics} lists the meeting topics that we use.

\begin{table*}[t]
\small
\centering
\begin{tabular}{@{}llll@{}}
\toprule
\textbf{Level 0} & \textbf{Level 1} & \textbf{Level 2} & \textbf{Level 3 (Examples)} \\
\midrule
\multirow{18}{*}{Non-Business} 
  & \multirow{5}{*}{Food} 
    & Cooking & Grilling techniques shared (\emph{... 9 more}) \\
  & & Baking & Bread making adventures (\emph{... 9 more}) \\
  & & Dining Out & Restaurant service reviews (\emph{... 9 more}) \\
  & & Food Culture & Cultural food traditions (\emph{... 9 more}) \\
  & & Nutrition & Vitamin supplement discussions (\emph{... 9 more}) \\
  & & \multicolumn{2}{l}{\emph{... 5 more subcategories}} \\
\cmidrule{2-4}
  & \multirow{5}{*}{Travel} 
    & Destinations & Mountain hiking trails (\emph{... 9 more}) \\
  & & Accommodations & Hotel loyalty programs (\emph{... 9 more}) \\
  & & Activities & Adventure sports planning (\emph{... 9 more}) \\
  & & Transportation Methods & Train journey experiences (\emph{... 9 more}) \\
  & & Travel Planning & Itinerary creation process (\emph{... 9 more}) \\
  & & \multicolumn{2}{l}{\emph{... 5 more subcategories}} \\
\cmidrule{2-4}
  & \multirow{5}{*}{Transportation} 
    & Public Transit & Bus schedule reliability (\emph{... 9 more}) \\
  & & Personal Vehicles & Car model comparisons (\emph{... 9 more}) \\
  & & Alternative Transport & Electric scooter trials (\emph{... 9 more}) \\
  & & Commuting & Route optimization strategies (\emph{... 9 more}) \\
  & & Ride Sharing & App comparison reviews (\emph{... 9 more}) \\
  & & \multicolumn{2}{l}{\emph{... 5 more subcategories}} \\
\midrule
\multirow{18}{*}{Business} 
  & \multirow{5}{*}{Drug Development} 
    & Preclinical Research & Animal Model Selection (\emph{... 9 more}) \\
  & & Clinical Trial Planning & Protocol Design Elements (\emph{... 9 more}) \\
  & & Regulatory Strategy & Submission Pathway Selection (\emph{... 9 more}) \\
  & & Partnership Collaborations & Due Process (\emph{... 9 more}) \\
  & & Biomarker Identification & Biomarker Discovery Methods (\emph{... 9 more}) \\
  & & \multicolumn{2}{l}{\emph{... 5 more subcategories}} \\
\cmidrule{2-4}
  & \multirow{5}{*}{Drug Approval} 
    & Regulatory Submission Preparation & Module Compilation Process (\emph{... 9 more}) \\
  & & FDA Communication & Meeting Request Preparation (\emph{... 9 more}) \\
  & & Clinical Data Analysis & Statistical Trial Evaluation (\emph{... 9 more}) \\
  & & Safety Reporting & Adverse Event Classification (\emph{... 9 more}) \\
  & & Manufacturing Compliance & GMP Inspection Preparation (\emph{... 9 more}) \\
  & & \multicolumn{2}{l}{\emph{... 5 more subcategories}} \\
\cmidrule{2-4}
  & \multirow{5}{*}{Drug Sales} 
    & Market Access Strategy & Payer Landscape Analysis (\emph{... 9 more}) \\
  & & Pricing Negotiations & Price Justification Strategy (\emph{... 9 more}) \\
  & & Sales Force Training & Product Knowledge Development (\emph{... 9 more}) \\
  & & Key Opinion Leader Engagement & KOL Identification Strategy (\emph{... 9 more}) \\
  & & Competitive Intelligence & Market Landscape Analysis (\emph{... 9 more}) \\
  & & \multicolumn{2}{l}{\emph{... 5 more subcategories}} \\
\bottomrule
\end{tabular}
\caption{Example non-meeting topic hierarchy for Biocure showing the four-level structure from top-level categories down to specific discussion topics.}
\label{fig:biocure-topics}
\end{table*}

\begin{table*}[t]
\centering
\small
\begin{tabular}{@{}lp{0.75\textwidth}@{}}
\toprule
\textbf{Meeting Type} & \textbf{Topics} \\
\midrule
Direct Report & Employee Performance Review, Goal Setting and Progress Check-In, Career Development Discussion, Feedback and Coaching Session, Workload and Prioritization Review, Upcoming Deadlines and Deliverables, Team Dynamics and Collaboration Feedback, Skill Growth and Training Opportunities, Manager-Report Relationship Check-In, Quarterly Performance Summary \\
\addlinespace
Team & Weekly Team Standup, Team Goals and KPIs Review, Process Improvement Discussion, Cross-Functional Collaboration Update, Retrospective and Lessons Learned, Upcoming Projects and Assignments, Team Communication and Workflow Alignment, Resource Planning and Allocation, Team Morale and Culture Check, Problem-Solving and Roadblock Discussion \\
\addlinespace
Task & Task Progress and Status Update, Dependency and Blocker Resolution, Timeline and Deliverable Planning, Quality Review and Standards Alignment, Task Ownership and Accountability, Cross-Task Integration Discussion, Testing and Validation Checkpoint, Task Documentation and Handover, Risk and Issue Review, Next Steps and Action Items \\
\addlinespace
Executive & Company Strategy and Vision Alignment, Financial Performance and Budget Review, Key Metrics and KPIs Discussion, Organizational Priorities and Focus Areas, Market Trends and Competitive Analysis, Resource Allocation and Headcount Planning, Operational Efficiency Review, Major Risks and Mitigation Planning, Executive Decisions and Approvals, Quarterly Business Review \\
\addlinespace
Project & Project Kickoff and Scope Definition, Milestone Progress Review, Timeline and Deliverable Alignment, Resource and Budget Status, Stakeholder Communication Plan, Risk Management and Mitigation Discussion, Project Dependencies and Coordination, Quality Assurance and Testing Review, Project Retrospective and Lessons Learned, Next Phase Planning and Execution \\
\addlinespace
Department & Department Strategy and Goals Review, Cross-Team Coordination Update, Department Resource Planning, Organizational Changes and Updates, Department Performance Metrics Review, Interdepartmental Collaboration Discussion, Budget and Resource Allocation Review, Department Culture and Team Building, Major Initiative Planning, Department Communication and Updates \\
\bottomrule
\end{tabular}
\caption{Meeting topics.}
\label{table:meeting-topics}
\end{table*}

\section{Meetings}
\label{app:meetings}
We simulate meetings for employees based on their organizational roles and relationships. \autoref{tab:meeting_types} defines the six meeting types supported in our simulation framework. Each meeting type is characterized by several parameters: a description of the participants, the probability of generating meeting minutes, the probability of generating an agenda, the probability of leaving the calendar event empty (without minutes or agenda), the meeting frequency, the designated organizer, and the knowledge graph relationships that the meeting substantiates.

For each meeting type, we determine participants from relevant organizational relationships. For example, direct report meetings involve an employee and their manager, while team meetings include all employees who belong to or lead a particular team. Meetings are scheduled according to their specified frequencies (weekly, biweekly, or monthly), and associated artifacts (meeting minutes and agendas) are generated based on the probabilities shown in \autoref{tab:meeting_types}. The "Empty Cal." column indicates the probability that a meeting appears on calendars without any accompanying documentation.

The "Substantiates" column indicates which knowledge graph relationships are evidenced or reinforced by each meeting type. For instance, direct report meetings substantiate task progress, reporting relationships, and work assignments, while executive meetings provide evidence of project progress and employee titles. This ensures that meetings generate realistic traces of organizational activity that align with the underlying knowledge graph structure.

This simulation process results in each employee having an average of 6 meetings per quarter, with individual counts varying based on their position, team membership, project involvement, and reporting relationships.

\begin{table*}[h!]
\centering
\footnotesize
\begin{tabular}{lp{1.8cm}ccccp{1.5cm}p{2.3cm}}
\toprule
\textbf{Meeting Type} & \textbf{Description} & \textbf{Minutes} & \textbf{Agenda} & \textbf{Empty Cal.} & \textbf{Frequency} & \textbf{Organizer} & \textbf{Substantiates} \\
\midrule
direct\_report & employee and their manager & 30\% & 30\% & 40\% & weekly & manager & task progress, reports to, works on \\
\midrule
team & employees that belong and lead a team & 25\% & 25\% & 50\% & monthly & team lead & task progress, works on, member of \\
\midrule
task & employees collaborating on a task & 30\% & 30\% & 40\% & biweekly & any & task progress, works on \\
\midrule
executive & employees that are executives & 50\% & 50\% & 0\% & monthly & any & project progress, title \\
\midrule
project & employees that work on tasks part of a project & 50\% & 50\% & 0\% & monthly & team lead of the project & project progress, works on, has task \\
\midrule
department & employees that lead the department and the teams part of it & 50\% & 50\% & 0\% & monthly & department lead & has task, team belongs to department, department belongs to company \\
\bottomrule
\end{tabular}
\caption{Supported Meeting Types.}
\label{tab:meeting_types}
\end{table*}

\section{ETL Pipeline}
\label{app:etl-process}

\paragraph{ETL for KB-Gold.}
To extract ground truth from knowledge base files into postgres databases accessible to our agent, we use the StoryBeat KB API to access entities, relationships, and documents.

\textbf{Extract}: 
We use the StoryBeat KB API's entity, relationship, and document interfaces to extract data from .kb files. 
The extraction uses generator functions to process data in configurable batches (default: 1000 items) rather than loading entire datasets into memory.

\textbf{Transform}: 
Entities are mapped from KB properties to database columns based on entity type (Company, Employee, Department, etc.). Each entity type has specific property mappings (e.g., Employee uses \texttt{hasName} for name, \texttt{hasAlias} for aliases).
Relationships are extracted with their temporal information (\texttt{beginning}, \texttt{end}) and classified as "ongoing" or "bounded".
Documents are processed separately using the KB documents API, with properties including sender/recipient information, dates, topics, and meeting references.
Array fields (aliases, recipient lists) are converted from string representations to PostgreSQL array types using \texttt{ast.literal\_eval}.

\textbf{Load}: 
Data is loaded into PostgreSQL using batch INSERT operations within transactions for consistency.
Entity tables use the KB entity ID as primary key (e.g., \texttt{employee\_EMP-D44FE8C5}).
The pipeline supports multiple schemas (one per company/knowledge base), with schema names inferred from KB filenames.
Duplicate detection uses \texttt{ON CONFLICT DO NOTHING} to handle idempotent loading.

\paragraph{ETL for RAG.} The RAG pipeline builds vector embeddings for document retrieval.

\textbf{Extract}: 
Documents are exported from KB files to a temporary directory using the KB documents API.

\textbf{Transform}:
Document content is truncated to 22,000 characters ($\approx$8,000 tokens) for embedding generation to stay within model limits. Embeddings are generated using OpenAI's text-embedding-3-small model (1536 dimensions) with parallel processing (4 workers by default).

\textbf{Load}:
Embeddings are stored in PostgreSQL using the pgvector extension with HNSW indexing for efficient similarity search.
Batch inserts (500 records per batch) optimize database write performance.
Duplicate documents are detected and skipped to avoid redundant API calls.
Full document content (not truncated) is stored alongside embeddings for retrieval.

\section{QA Benchmarks}
\label{app:qa-benchmarks}

\lstset{
  basicstyle=\small\ttfamily,
  breaklines=true,
  breakatwhitespace=true,
  columns=flexible,
  keepspaces=true,
  breakindent=0pt,
  resetmargins=true,
}

\newcommand{\qatable}[6][]{\begin{table}[#1]
\centering
\small
\begin{tabular}{@{}l|p{0.65\linewidth}@{}}
    \toprule
    \textbf{Question} & #2 \\
    \textbf{Answer} & #3 \\
    \textbf{Company/KB} & #4 \\
    \textbf{Answer Type} & #5 \\
    \textbf{Constraints} & #6 \\
    \bottomrule
\end{tabular}
\end{table}
}

In the following section, we outline sample questions from three QA dataset types across the different companies they are generated from. In each example, we include the minimal set of documents required to review/analyze to be able to fully answer the presented question. There may be multiple minimal sets of documents to answer said questions, of which we select one. The selected questions are a small subset of all questions in the dataset and aim to show the diversity of questions, answers, and documents. Many questions require analyzing many documents (e.g. "Who are the employees of pound financial group llc?"), but these questions are omitted for the sake of conciseness.

For each question, we measure the number of constraints, which represents the number of filters on the documents or KB that an agent must apply to answer a question. However, this doesn't necessarily correlate with question difficulty, because low constraint count questions may be more difficult to answer due to their expansive breadth (see example in paragraph above).

\section{KB QA}
\label{app:kbqa}
\qatable[h!]{Who participated in the meeting titled cross-team protocol alignment - work session?}{James Moreira, James Vera}{Streamvibe}{List[str]}{1}

\begin{simplegray}{calendar\_invite\_james\_vera\_f685.txt}
\begin{lstlisting}[escapechar=!]
Subject: !\textbf{Cross-team protocol alignment - Work Session}!

From: James Vera <vera.Jamie@streamvibestudios.com>
To: !\textbf{James Vera}! <vera.Jamie@streamvibestudios.com>, !\textbf{James Moreira}! <Jim.moreira@streamvibestudios.com>
Date: 2024-02-22

CALENDAR INVITE

You are invited to: !\textbf{Cross-team protocol alignment - Work Session}!
Scheduled for: 2024-02-22

Organizer: James Vera (vera.Jamie@streamvibestudios.com)

Attendees:
  - !\textbf{James Vera}! (vera.Jamie@streamvibestudios.com)
  - !\textbf{James Moreira}! (Jim.moreira@streamvibestudios.com)

This meeting has been scheduled by James Vera.

Please accept or decline this invitation at your earliest convenience.
\end{lstlisting}
\end{simplegray}

\noindent
The calendar invite clearly lists out all attendees: \verb|James Vera|, \verb|James Moriera|. We judge these responses as sets, so ordering doesn't matter.

\qatable[h!]{Who organized the target company due diligence discussion meeting?}{Birgitta Lagarde}{Pound}{str}{1}

\begin{simplegray}{meeting\_minute\_Quality\\\_Review\_and\_Standards\_Alignment\_07a7.txt}
\begin{lstlisting}[escapechar=!]
Message-ID: 61f61ba7-8cd2-4a17-a311-6cdfd059f7bc
From: !\textbf{Birgitta Lagarde}! <blagarde@poundfinancialgroupllc.com>
To: Piermaria Ferguson <pferguson@poundfinancialgroupllc.com>
Date: 2024-01-03
Subject: !\textbf{Target company due diligence Discussion}!
MIME-Version: 1.0
Content-Type: text/plain; charset=UTF-8
Content-Transfer-Encoding: 7bit

Piermaria,

Thanks for making time to sync up on the !\textbf{target company due diligence}! work we've got going. I wanted to recap what we covered in our last meeting and make sure we're both tracking the same priorities as we move forward with this.

We spent some good time going through our quality review process and making sure our standards are aligned across the board. We talked through the key documentation we need to pull together, identified some gaps in our initial assessment, and went over what the next phase of verification looks like. It was helpful to lock in on our approach and get clear on who's handling what pieces.

Here's where things stand with our progress:

!\textbf{Target company due diligence}! is advancing steadily with you and I, around 30-40% finished.

I think we're in a solid spot to keep pushing forward on this. Let's keep the momentum going and touch base again soon on any blockers or adjustments we need to make.

Thank you,
!\textbf{Birgitta}!

!\textbf{Birgitta Lagarde}!
Analyst
Pound Financial Group LLC

blagarde@poundfinancialgroupllc.com
Desk: +1 (650) 925-3434
Mobile: +1 (650) 205-2167
[INSERT_BOX_LOGO]
\end{lstlisting}
\end{simplegray}

\noindent
The meeting minutes document implicitly indicates that \verb|Birgitta Lagarde|, as the organizer, is sending meeting minutes to the sole other attendee Piermaria Ferguson.

\section{Topic QA}

\qatable[h!]{When was drug approval first mentioned?}{2024-01-01}{Biocure}{date}{1}

\begin{simplegray}{email\_Label\_Negotiation\_Process\_546e.txt}
\begin{lstlisting}[escapechar=!]
Message-ID: 3ce33e9f-3d3c-4100-bf30-b2fc26cbbd05
From: Jeronimo Zobel <jeronimo.zobel@yahoo.com>
To: Manuella Barrios <manuella@gmail.com>
Date: !\textbf{2024-01-01}!
Subject: Update on !\textbf{drug approval}! labeling coordination
MIME-Version: 1.0
Content-Type: text/plain; charset=UTF-8
Content-Transfer-Encoding: 7bit

Hi Manuella, I wanted to touch base on a couple of items related to our business operations and the !\textbf{drug approval}! process. As we continue managing our labeling requirements across various products, I've been thinking through how we can streamline our approach to the label negotiation process with regulatory bodies. The negotiations have been progressing, but I believe we could benefit from a more structured timeline and clearer documentation of our discussions with the FDA.

Recently, manuella, checking in with you as my direct supervisor, and I wanted to ensure we're aligned on priorities for the labeling requirements phase. The label negotiation process is becoming more complex as we handle multiple products simultaneously, and having direct oversight will help us maintain consistency across submissions. I think it would be valuable to establish checkpoints where we review our negotiation strategy and any feedback we've received from regulatory reviewers.

I'd like to schedule a brief meeting this week to discuss our current bottlenecks in the !\textbf{drug approval}! process, particularly around the labeling timeline. Once we optimize the label negotiation process, I believe we can accelerate our overall business operations in this area. Let me know your availability and if there are specific concerns you'd like to address.

Thank you,
Jeronimo Zobel

Jeronimo Zobel
Systems Administrator
+1 (650) 682-8707 | jeronimo.z@biocuresolutions.com
\end{lstlisting}
\end{simplegray}

\noindent
The email is the first mention of \verb|Drug Approval|. The topic hierarchy for this email is as follows: (0) \verb|Label Negotiation Process|, (1) \verb|Labeling Requirements|, and (2) \verb|Drug Approval|. See \autoref{app:topic-levels} for a more detailed explanation on topic levels. The higher-level topics are interwoven into the main content of the email, which is centered around the more granular topic of \verb|Label Negotiation Process|.

\section{Integrated QA}

\qatable[h!]{Did anyone who works at Zenith Labs send emails about Daily commute issues during month 1 of 2024?}{true}{Zenith}{bool}{4}

\begin{simplegray}{email\_Daily\_commute\_issues\_5201.txt}
\begin{lstlisting}[escapechar=§]
Message-ID: f57b61b0-4543-423d-9172-e8f1753abb7f
From: Albert Garcia §\textbf{<Bert.garcia@zenithlabs.com>}§
To: Emilia Anglada <eanglada@hotmail.com>
Date: 2024-01-10
Subject: §\textbf{Morning commute struggles}§
MIME-Version: 1.0
Content-Type: text/plain; charset=UTF-8
Content-Transfer-Encoding: 7bit

Hi Emilia,

I wanted to check in since we've both been navigating some rough commute situations lately. The traffic on my usual route has been absolutely brutal the past couple of weeks, and I've been leaving earlier just to make it to the office on time. I know you mentioned something similar last week when we were chatting by the desks.

I've been thinking about how these §\textbf{daily commute issues}§ really affect our whole day and energy levels at work. When you're stuck in traffic or dealing with transit delays, it's hard to come in feeling fresh and ready to tackle complex projects. Building on that, accepted the firmware performance benchmarking role this morning, so I've been reflecting on how important it is to stay focused even when the morning doesn't go as planned.

I've actually started trying out a different route a couple times this week, and it's made a noticeable difference on days when the main highway is congested. Leaving just fifteen minutes earlier has been a game-changer for me, and it gives me time to grab coffee and settle in before diving into emails. Maybe we could compare notes on what's been working for each of us?

Anyway, just wanted to reach out and see if you've found any good solutions to the commute chaos. It seems like a lot of us in the office have been dealing with similar frustrations, so I figured it might be worth discussing what strategies have helped. Hope things smooth out for you soon!

Thanks,
Albert Garcia
Marketing Specialist, Sales & Marketing
§\textbf{Zenith Labs}§

+1 (415) 400-8511 | Bert.garcia@zenithlabs.com
[INSERT_HORIZONTAL_LOGO]
\end{lstlisting}
\end{simplegray}

\noindent
In this Email, the sender \verb|Albert Garcia| can be identified as a Zenith Labs employee by his signature or his email domain. The topic of the email is \verb|Daily commute issues| and it was sent on \verb|2024-01-10|, which falls in the date range of the question.

\qatable[h!]{How many of jared chartier's direct reports sent emails about technical debt management after 29 Mar 2024?}{0}{Zenith}{int}{3}

\begin{simplegray}{email\_Technical\_Debt\_Management\_72dc.txt}
\begin{lstlisting}[escapechar=§]
Message-ID: 54b67748-00c1-4d5c-b996-62c9e0241545
From: §\textbf{Emilia Anglada}§ <eanglada@hotmail.com>
To: §\textbf{Jared Chartier}§ <jaredchartier@gmail.com>
Date: §\textbf{2024-01-10}§
Subject: Addressing code quality concerns in our operations
MIME-Version: 1.0
Content-Type: text/plain; charset=UTF-8
Content-Transfer-Encoding: 7bit

Hi §\textbf{Jared}§, I've been reflecting on how our current business operations could benefit from a more intentional approach to §\textbf{managing technical debt}§. As our codebase continues to grow, I'm noticing that we're accumulating layers of complexity that might slow down our team's ability to iterate efficiently. I think it would be valuable for us to establish a clearer strategy around this.

Building on that concern, §\textbf{Jared, I wanted to get your direction here}§ on how we should prioritize which areas to tackle first. I'm thinking we could start by identifying the components that have the highest maintenance burden and then work through a systematic refactoring plan. Would you be open to discussing this further so we can align on the best path forward for our technical infrastructure?

Best,
§\textbf{Emilia Anglada}§
Marketing Specialist
+1 (650) 524-2805
\end{lstlisting}
\end{simplegray}

\noindent
This is the only email about \verb|Technical Debt Management| that a direct report of \verb|Jared Chartier| has sent. Due to the size of the Zenith Labs company, \verb|Emilia Anglada| is also the only direct report. The email was sent on January 10, 2024, so it doesn't fall within the date range of the question. Also, the mention of \verb|"I want to get your direction here"| serves as an implicit mention of Emilia reporting to Jared.

\section{QA Dataset Assumptions}
\label{app:qa-dataset_assumptions}

We explicitly state the following assumptions and details we were using to create our QA ensure clarity and proper use:
\begin{itemize}
    \item Ground truth answers are obtained through manually-verified SPARQL queries executed on the underlying knowledge base. This verification process ensures that the correct answers are returned regardless of which template variables are instantiated.
    \item All questions requesting lists (e.g., "Who are all the employees of X?") return results without duplicate entries. This deduplication is applied consistently across the dataset.
    \item Questions exclusively target regular email topics and intentionally exclude meeting-related topics, as these constitute distinct semantic categories with different structural properties.
    \item Due to the scale of the Pound knowledge base, minimally-constrained questions often yield extremely large answer sets. For example, the query "Who is an employee of Pound?" returns several thousand names, potentially exceeding typical LLM context window limits. In contrast, questions with additional constraints produce more manageable, specific answer sets that are better suited for LLM agent processing.
    \item The answer key generation process applies SELECT DISTINCT operations to remove duplicate names from result sets. Consequently, a minor discrepancy may exist between count-based answers and the actual length of returned name lists, as overlapping names are consolidated into single entries.
\end{itemize}

\section{Experimental Setup}
\label{app:experimental_setup}

\paragraph{KB Evaluation}

\begin{itemize}[itemsep=-5pt]
    \item \textbf{Datasets:} All documents from the four company datasets from \autoref{tab:company_datasets}.
    
    \item \textbf{Models:} Claude Haiku 4.5, GPT-5 Nano, Gemini 2.5 Flash Lite.
    
    \item \textbf{Prompting:} Few-shot with 8 examples (prompt template in \autoref{app:extraction-prompt}). We selected samples to properly cover diverse entity/relationship types when providing examples.
    
    \item \textbf{Entity Evaluation:} Per-type precision, recall, and $F_1$, macro-averaged across types. TP: matched entities; FP: spurious extractions; FN: missed ground truth entities.
    
    \item \textbf{Relationship Evaluation:} Dual evaluation measuring (1) structural correctness (triple exists) and (2) temporal accuracy (triple with correct dates). Per-type metrics macro-averaged across relationship types. Temporal evaluation requires both correct triple structure and matching start/end dates.
    
    \item \textbf{Implementation:} Temperature=1.0, max tokens=8,192, exponential backoff retry (max 3 attempts, 60s timeout).
\end{itemize}

\paragraph{Topic Classification}

\begin{itemize}[itemsep=-5pt]
    \item \textbf{Datasets:} Biocure (L, 31 topics) and Pound (XL, 17 topics) from \autoref{tab:company_datasets}. Topic definitions in \autoref{tab:topic-categories}.
    
    \item \textbf{Models:} Claude Haiku 4.5, Claude Sonnet 4.5, GPT-5 Nano, GPT-5.1, and Gemini 2.5 Flash Lite.

    \item \textbf{Baseline:} TF-IDF features with Logistic Regression classifier.
    
    \item \textbf{Train/Test Split:} Stratified sampling with $k \in \{0, 10, 50\}$ training examples for LLMs and $k \in \{10, 50, 100, 1000, 10000, 100000\}$ for baseline. Test set: 1,000 examples per configuration (or maximum available). 5 repetitions with different random splits.
    
    \item \textbf{Prompting:} Zero-shot ($k=0$) or few-shot ($k>0$) examples per topic (template in \autoref{app:topic-prompt}).
    
    \item \textbf{Evaluation:} Per-topic precision, recall, and $F_1$, macro-averaged across topics. 
    
    \item \textbf{Implementation:} Temperature=1.0, max tokens=500, exponential backoff retry (max 3 attempts, 60s timeout).
\end{itemize}

\paragraph{QA}
\begin{itemize}[itemsep=-5pt]
    \item \textbf{Datasets}: KB QA, Topic QA, and Integrated QA. 250 Questions * 4 KB sizes * 3 QA Datasets = 3k questions. Single full dataset run.
    \item \textbf{Models}: Claude Haiku 4.5, Claude Sonnet 4.5, GPT-5 Nano, GPT-5.1, and Gemini 2.5 Flash Lite. Used through Anthropic API, OpenAI API, and OpenRouter respectively.
    \item \textbf{Methods}: RAG ($k=10$): Agent with access to a tool to query a vector db. KB: Agent with access to a tool to query a Postgres DB with SQL. DBs contain ground truth information extracted from the KB and follow a standard relational schema (not relevant for RAG Agent). Database access is restricted for each dataset. Restriction applies based on KB, method, and dataset to ensure no data leakage happens: e.g. when answering questions in Topic QA for Biocure, the KB agent will exclusively have access to the table that stores information on the documents from Biocure.
    \item \textbf{Evaluation}: Case-insensitive exact match (string, date, integer, boolean) scoring 0/1 and $F_1$ score for set[string] questions to allow models to obtain partial credits (as these questions are typically the hardest to answer). For boolean we accepted alternative wording such as ``Yes'' and ``No''.
    \item \textbf{Implementation}: Pydantic AI agent with custom tools to retrieve from Vector DB (RAG) or Postgres DB (KB). Tool call limit of 5. Few shot examples (specific to each QA dataset) were provided for each prompt.
\end{itemize}

\section{QA Results}
\label{app:experimental-results}
\begin{figure*}
    \centering
    \includegraphics[width=\textwidth]{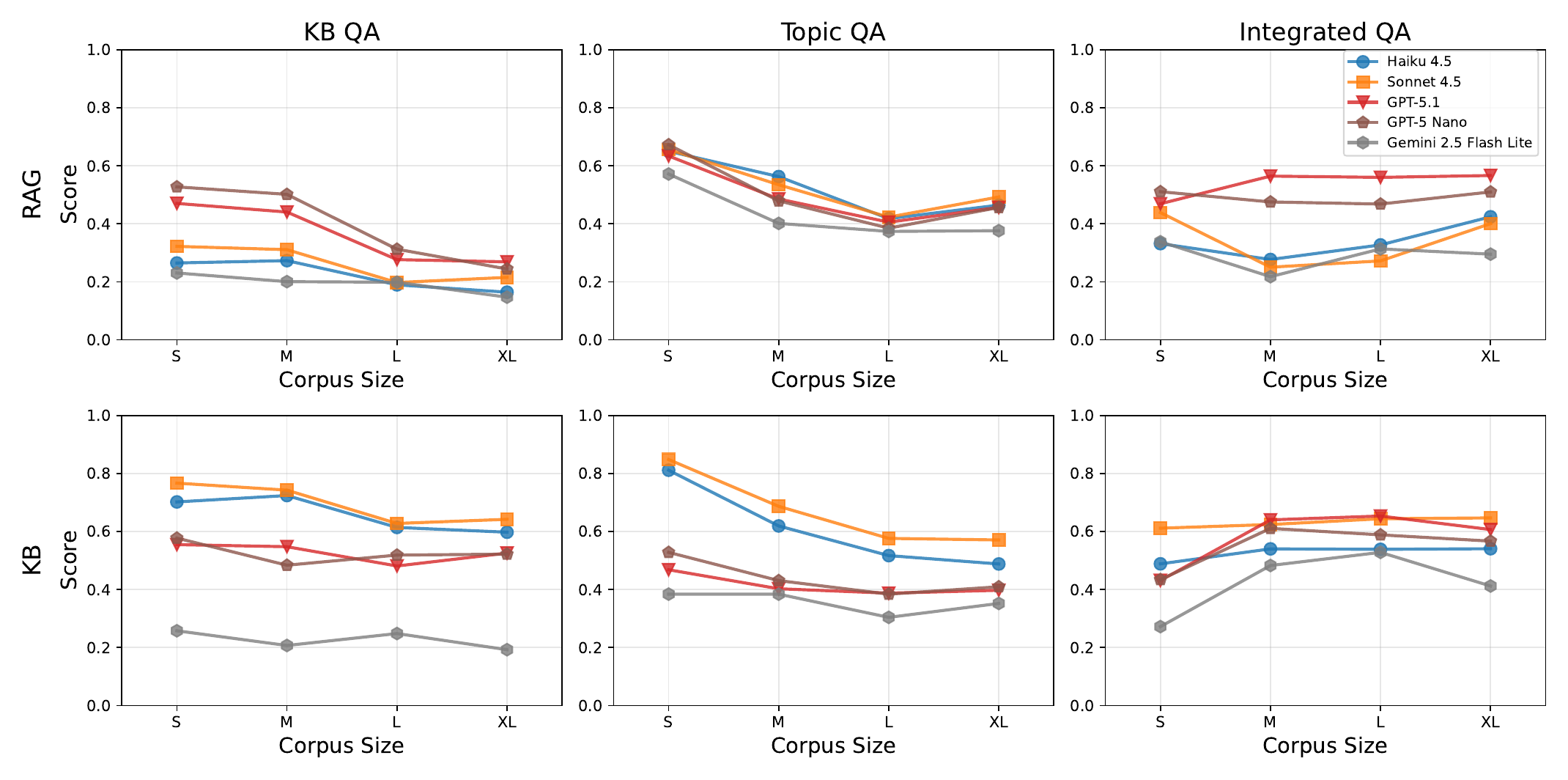}
    \caption{Visualization of results presented in \autoref{tab:qa-results}.}
    \label{fig:qa_results_by_model}
\end{figure*}

\begin{figure*}
    \centering
    \includegraphics[width=\textwidth]{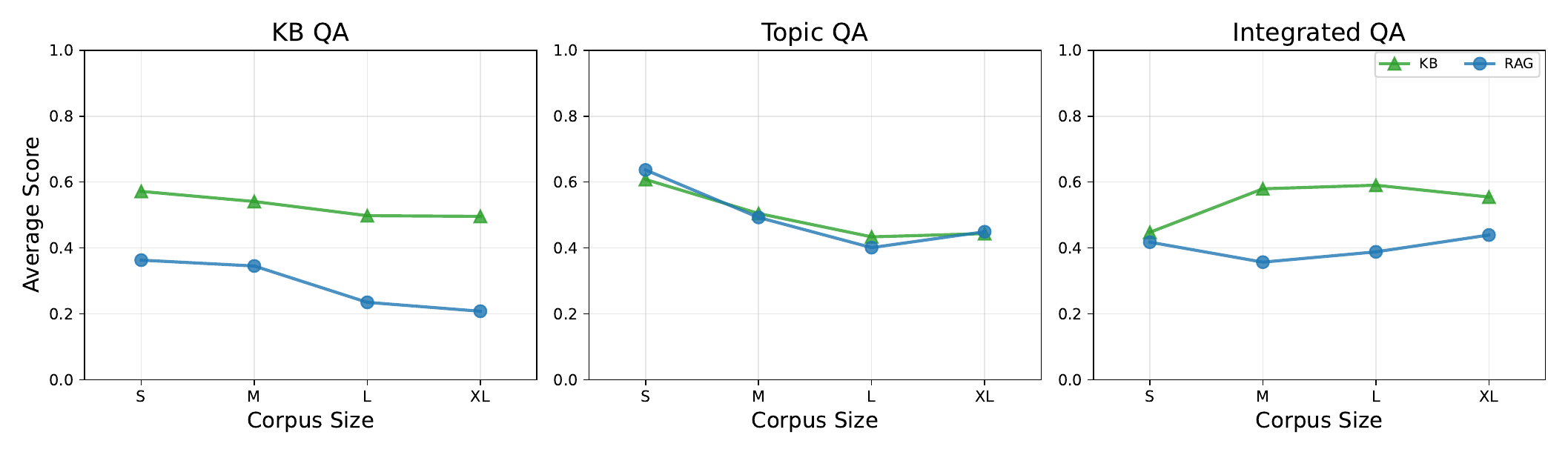}
    \caption{Visualization of results presented in \autoref{tab:qa-results} aggregated by method.}
    \label{fig:qa_results_by_method}
\end{figure*}

In this section, we visualize the results we have for QA in graphs. \autoref{fig:qa_results_by_model} shows the QA results by model and \autoref{fig:qa_results_by_method} shows the QA results by method.

\lstset{
    breaklines=true,
    basicstyle=\small\ttfamily,
    breakatwhitespace=false,
    postbreak=\mbox{\textcolor{red}{$\hookrightarrow$}\space} }

\section{KB Evaluation Results}
\label{app:kb-eval-results}

Table~\ref{tab:error_distribution} reveals distinct error patterns across relationship types and models. For non-temporal relationships, Gemini 2.5 Flash Lite exhibits the highest over-prediction rate for worksOn (69\%), while all models show substantial under-prediction of attend relationships (63–71\% FN rates). Temporal relationship extraction proves more challenging: Gemini 2.5 Flash Lite over-predicts worksOn at 56\%, while Haiku 4.5 shows notable over-prediction of attend (45\%) and organize (25\%). Under-prediction of attend remains problematic across temporal contexts (50–54\% FN rates for all models). These patterns suggest that models struggle with event-based relationships (attend, organize) regardless of temporal context, while structural relationships (worksOn, memberOf) are prone to over-prediction, particularly for Gemini 2.5 Flash Lite. The relatively balanced error rates for core relationships like reportsTo and worksAt (mostly < 10\%) indicate better model performance for hierarchical organizational structures.

\begin{table*}[htbp]
\centering
\small
\begin{tabular}{llccc}
\toprule
\textbf{Error Type} & \textbf{Relationship} & \textbf{Gemini 2.5 FL} & \textbf{Haiku 4.5} & \textbf{GPT-5 Nano} \\
\midrule
\multicolumn{5}{l}{\textit{Non-Temporal Relationships}} \\
\midrule
\textbf{Over-predicts (FP)} & worksOn & 69\% & 64\% & 43\% \\
& worksAt & 7\% & $-$ & 3\% \\
& reportsTo & 8\% & $-$ & 4\% \\
& memberOf & 16\% & 12\% & 50\% \\
& attend & $-$ & 16\% & $-$ \\
& organize & $-$ & 8\% & $-$ \\
\midrule
\textbf{Under-predicts (FN)} & worksOn & 8\% & 15\% & 16\% \\
& worksAt & 5\% & $-$ & $-$ \\
& reportsTo & 1\% & 6\% & 4\% \\
& memberOf & 3\% & 3\% & 3\% \\
& attend & 68\% & 71\% & 63\% \\
& organize & 15\% & 5\% & 14\% \\
\midrule
\multicolumn{5}{l}{\textit{Temporal Relationships}} \\
\midrule
\textbf{Over-predicts (FP)} & worksOn & 56\% & 23\% & 43\% \\
& worksAt & 4\% & 1\% & 16\% \\
& reportsTo & 21\% & $-$ & 4\% \\
& memberOf & 19\% & 6\% & 37\% \\
& attend & $-$ & 45\% & $-$ \\
& organize & $-$ & 25\% & $-$ \\
\midrule
\textbf{Under-predicts (FN)} & worksOn & 31\% & 34\% & 36\% \\
& memberOf & 3\% & 3\% & 3\% \\
& attend & 54\% & 52\% & 50\% \\
& organize & 12\% & 11\% & 11\% \\
\bottomrule
\end{tabular}
\caption{Error Distribution by Relationship Type and Model}
\label{tab:error_distribution}
\vspace{4pt}
\end{table*}

\section{Topic Classification Results}
\label{app:topic-classification-results}

\begin{table*}[htbp]
\centering
\small
\begin{tabular}{lrcccccc}
\toprule
\multirow{2}{*}{\textbf{Model}} & \multirow{2}{*}{\textbf{Training Size}} & \multicolumn{3}{c}{\textbf{Biocure (L)}} & \multicolumn{3}{c}{\textbf{Pound (XL)}} \\
\cmidrule(lr){3-5} \cmidrule(lr){6-8}
 &  & \textbf{Precision} & \textbf{Recall} & \textbf{F$_1$} & \textbf{Precision} & \textbf{Recall} & \textbf{F$_1$} \\
\midrule

\multirow{3}{*}{Haiku 4.5}
 & 0 & .917$\pm$.00 & .887$\pm$.01 & .885$\pm$.01 & .550$\pm$.01 & .410$\pm$.02 & .357$\pm$.01 \\

 & 10 & .950$\pm$.01 & .959$\pm$.01 & .951$\pm$.01 & .649$\pm$.01 & .650$\pm$.01 & .594$\pm$.01 \\

 & 50 & .963$\pm$.01 & .967$\pm$.00 & \textbf{.963$\pm$.01} & .697$\pm$.01 & .733$\pm$.01 & \textbf{.685$\pm$.01} \\
\midrule
\multirow{3}{*}{Sonnet 4.5}
 & 0 & .875$\pm$.04 & .873$\pm$.04 & .872$\pm$.04 & .583$\pm$.01 & .544$\pm$.02 & .517$\pm$.02 \\

 & 10 & .963$\pm$.00 & .966$\pm$.01 & .962$\pm$.01 & .681$\pm$.03 & .692$\pm$.03 & .661$\pm$.03 \\

 & 50 & .979$\pm$.00 & .982$\pm$.00 & \textbf{.980$\pm$.00} & .748$\pm$.02 & .777$\pm$.02 & \textbf{.751$\pm$.02} \\
\midrule
\multirow{3}{*}{GPT-5 Nano}
 & 0 & .905$\pm$.00 & .901$\pm$.01 & \textbf{.898$\pm$.01} & .626$\pm$.00 & .563$\pm$.01 & .531$\pm$.01 \\

 & 10 & .897$\pm$.01 & .898$\pm$.02 & .893$\pm$.02 & .661$\pm$.01 & .583$\pm$.01 & .585$\pm$.00 \\

 & 50 & .878$\pm$.03 & .884$\pm$.03 & .877$\pm$.03 & .642$\pm$.01 & .615$\pm$.01 & \textbf{.598$\pm$.01} \\
\midrule
\multirow{3}{*}{GPT-5.1}
 & 0 & .954$\pm$.02 & .952$\pm$.03 & .950$\pm$.03 & .620$\pm$.01 & .524$\pm$.01 & .483$\pm$.00 \\

 & 10 & .975$\pm$.00 & .979$\pm$.00 & .977$\pm$.00 & .709$\pm$.02 & .721$\pm$.02 & .684$\pm$.02 \\

 & 50 & .981$\pm$.00 & .985$\pm$.00 & \textbf{.982$\pm$.00} & .762$\pm$.01 & .778$\pm$.01 & \textbf{.748$\pm$.02} \\
\midrule
\multirow{3}{*}{Gemini 2.5 F-L}
 & 0 & .938$\pm$.01 & .931$\pm$.01 & \textbf{.929$\pm$.01} & .563$\pm$.02 & .456$\pm$.02 & .412$\pm$.02 \\

 & 10 & .826$\pm$.01 & .841$\pm$.00 & .831$\pm$.01 & .609$\pm$.01 & .656$\pm$.02 & .620$\pm$.02 \\

 & 50 & .873$\pm$.01 & .889$\pm$.01 & .879$\pm$.01 & .669$\pm$.01 & .713$\pm$.01 & \textbf{.682$\pm$.01} \\
\midrule
\multirow{6}{*}{TF-IDF}
 & 10 & .036$\pm$.05 & .122$\pm$.11 & .052$\pm$.06 & .050$\pm$.07 & .101$\pm$.09 & .046$\pm$.08 \\

 & 50 & .169$\pm$.09 & .114$\pm$.08 & .090$\pm$.07 & .264$\pm$.02 & .279$\pm$.03 & .226$\pm$.03 \\

 & 100 & .379$\pm$.02 & .194$\pm$.02 & .206$\pm$.02 & .295$\pm$.01 & .388$\pm$.01 & .332$\pm$.01 \\

 & 1000 & .969$\pm$.02 & .934$\pm$.02 & .942$\pm$.02 & .900$\pm$.01 & .677$\pm$.02 & .709$\pm$.03 \\

 & 10000 & .994$\pm$.00 & .993$\pm$.00 & \textbf{.993$\pm$.00} & .970$\pm$.01 & .949$\pm$.01 & .958$\pm$.01 \\

 & 100000 & - & - & - & .986$\pm$.00 & .981$\pm$.00 & \textbf{.983$\pm$.00} \\

\bottomrule
\\[-3mm]
\multicolumn{8}{l}{\small $^{\ddagger}$ Gemini 2.5 \textbf{Flash Lite}.} \\
\end{tabular}
\caption{Topic Classification Results.}

\label{tab:topic-classification-results}
\end{table*}

Table~\ref{tab:topic-classification-results} presents topic classification performance across models and training sizes. On Biocure (L), all models achieve strong zero-shot performance (F1: 0.872–0.950), with minimal improvement in few-shot settings (50-shot F1: 0.877–0.982). The TF-IDF+LR baseline demonstrates clear data scaling, reaching F1 of 0.993 with 10K examples -- comparable to LLM zero-shot performance but requiring substantially more training data. On Pound (XL), LLMs show moderate zero-shot performance (F1: 0.357–0.531) with notable few-shot improvements (50-shot F1: 0.598–0.751). The data-rich baseline achieves superior performance at 100K examples (F1: 0.983), outperforming all LLMs.

\section{QA Error Analysis}
\label{app:qa-error-analysis}

In total, we ran 250 questions across 5 LLM models for 4 companies in 3 QA settings with 2 methods (RAG and KB). Thus, we had 30,000 total model runs. \autoref{tab:errors_consolidated} shows the programmatic errors that we ran into for each model in each setting (Topic QA, KB QA, and Integrated QA).

Our analysis reveals substantial variation in error rates across models, with GPT-5.1 experiencing the highest failure rate (247 errors, 4.1\% of its runs) and Sonnet 4.5 showing the most robust performance (91 errors, 3.0\% of its runs). The total error count across all models was 697 out of 30,000 runs, representing a 2.3\% overall failure rate.

KB-based queries produced 4.7× more errors than RAG-based queries (547 vs. 150 errors). This disparity suggests that knowledge base integration introduces additional failure modes, particularly related to tool calling and output validation. The complexity of KB queries—which require models to navigate structured data and execute multiple tool calls— contributes to this elevated error rate.

Error rates exhibit a clear positive correlation with query complexity (S→M→L→XL). For KB queries, XL-scale questions generated 175 errors compared to just 107 errors for S-scale questions. This pattern is particularly pronounced for context length issues: "Prompt too long" errors occur almost exclusively at L and XL scales, accounting for 72 such errors. This suggests that as queries grow in complexity, models struggle with context management and multi-step reasoning.

\begin{table*}[t]
\centering
\small
\begin{tabular}{@{}llccccccccc@{}}
\toprule
  &   & \multicolumn{4}{c}{\textbf{RAG}} & \multicolumn{4}{c}{\textbf{KB}} &   \\
\cmidrule(lr){3-6} \cmidrule(lr){7-10}
\textbf{Model} & \textbf{Error Type} & \textbf{S} & \textbf{M} & \textbf{L} & \textbf{XL} & \textbf{S} & \textbf{M} & \textbf{L} & \textbf{XL} & \textbf{Total} \\
\midrule
\multirow{4}{*}{Haiku 4.5} & Prompt too long & 0 & 0 & 0 & 2 & 0 & 0 & 8 & 22 & 32 \\
  & Tool calls limit exceeded & 0 & 0 & 0 & 0 & 19 & 15 & 26 & 24 & 84 \\
  & Tool max retries exceeded & 0 & 0 & 0 & 0 & 0 & 0 & 3 & 0 & 3 \\
\cmidrule(l){2-11}
  & \textbf{Total Errors} & \textbf{0} & \textbf{0} & \textbf{0} & \textbf{2} & \textbf{19} & \textbf{15} & \textbf{37} & \textbf{46} & \textbf{119} \\
\midrule
\multirow{6}{*}{Sonnet 4.5} & Output validation failed & 0 & 6 & 9 & 1 & 0 & 11 & 10 & 1 & 38 \\
  & Prompt too long & 0 & 0 & 0 & 2 & 0 & 0 & 2 & 36 & 40 \\
  & Request timeout & 0 & 0 & 0 & 0 & 0 & 0 & 3 & 2 & 5 \\
  & Tool calls limit exceeded & 0 & 0 & 0 & 0 & 0 & 1 & 3 & 3 & 7 \\
  & Tool max retries exceeded & 0 & 0 & 0 & 0 & 0 & 0 & 1 & 0 & 1 \\
\cmidrule(l){2-11}
  & \textbf{Total Errors} & \textbf{0} & \textbf{6} & \textbf{9} & \textbf{3} & \textbf{0} & \textbf{12} & \textbf{19} & \textbf{42} & \textbf{91} \\
\midrule
\multirow{6}{*}{GPT-5 Nano} & Empty model response & 2 & 2 & 1 & 4 & 3 & 13 & 7 & 0 & 32 \\
  & Other API errors & 0 & 0 & 0 & 6 & 0 & 1 & 0 & 2 & 9 \\
  & Request timeout & 0 & 0 & 0 & 0 & 0 & 0 & 0 & 1 & 1 \\
  & Server error (5xx) & 0 & 0 & 0 & 0 & 0 & 0 & 0 & 1 & 1 \\
  & Tool calls limit exceeded & 0 & 0 & 2 & 4 & 18 & 15 & 26 & 20 & 85 \\
\cmidrule(l){2-11}
  & \textbf{Total Errors} & \textbf{2} & \textbf{2} & \textbf{3} & \textbf{14} & \textbf{21} & \textbf{29} & \textbf{33} & \textbf{24} & \textbf{128} \\
\midrule
\multirow{7}{*}{GPT-5.1} & Empty model response & 0 & 0 & 19 & 26 & 39 & 36 & 69 & 23 & 212 \\
  & Other API errors & 0 & 0 & 0 & 0 & 0 & 0 & 0 & 3 & 3 \\
  & Rate limit exceeded & 0 & 6 & 4 & 0 & 4 & 11 & 0 & 0 & 25 \\
  & Server error (5xx) & 0 & 0 & 0 & 0 & 0 & 0 & 0 & 1 & 1 \\
  & Tool calls limit exceeded & 0 & 0 & 0 & 0 & 2 & 1 & 0 & 2 & 5 \\
  & Unknown error & 0 & 0 & 1 & 0 & 0 & 0 & 0 & 0 & 1 \\
\cmidrule(l){2-11}
  & \textbf{Total Errors} & \textbf{0} & \textbf{6} & \textbf{24} & \textbf{26} & \textbf{45} & \textbf{48} & \textbf{69} & \textbf{29} & \textbf{247} \\
\midrule
\multirow{5}{*}{Gemini 2.5 Flash Lite} & Invalid API response & 0 & 1 & 1 & 0 & 1 & 2 & 3 & 4 & 12 \\
  & Output validation failed & 2 & 1 & 2 & 0 & 19 & 23 & 10 & 25 & 82 \\
  & Tool calls limit exceeded & 0 & 0 & 0 & 0 & 2 & 5 & 5 & 5 & 17 \\
  & Unknown error & 0 & 0 & 0 & 0 & 0 & 1 & 0 & 0 & 1 \\
\cmidrule(l){2-11}
  & \textbf{Total Errors} & \textbf{2} & \textbf{2} & \textbf{3} & \textbf{0} & \textbf{22} & \textbf{31} & \textbf{18} & \textbf{34} & \textbf{112} \\
\bottomrule
\end{tabular}
\caption{Error analysis across all models.}
\label{tab:errors_consolidated}
\end{table*}

\clearpage
\section{KB Evaluation Prompt}
\label{app:extraction-prompt}

The following is the extraction prompt used for the KB Evaluation task.

\begin{lstlisting}
    
# Business Document Knowledge Extraction System

You are an expert knowledge extraction system. Your task is to analyze business documents and extract structured information about entities (people, organizations, projects, tasks, meetings) and the relationships between them.

## Output Requirements

**CRITICAL**: Return ONLY valid JSON with no explanations, preambles, or markdown formatting. The response must be parseable JSON that exactly matches the schema provided.

---

## Step 1: Document Classification

Before extracting any information, carefully read the entire document and classify it as ONE of the following types:

- **`agenda`**: An email outlining topics and agenda for an upcoming meeting
- **`meeting_minute`**: An email summarizing the discussions and outcomes of a past meeting
- **`regular_email`**: An email message between employees NOT discussing topics related to upcoming or past meetings

Store this classification as `document_type` in your response.

---

## Step 2: Entity Extraction

### Entity Extraction Rules

1. **Extract ONLY entities that participate in relationships** - no orphaned entities. For instance, if the relationships are "Employee worksOn Task" and "Employee worksAt Company" then ONLY extract entities Employee, Task, and Company, nothing else. Do the same for each relationship.
2. Extract entities exactly as they appear in the document (preserve exact wording for aliases)
3. For each entity, capture ALL variations/aliases found in the document (e.g., ["John Smith", "J. Smith", "John", "John S.", "Jonny"] etc.)
4. Entity aliases MUST be extracted exactly as they appear in the document (do not make up aliases)
5. Email addresses are CRITICAL for entity linking - extract when available
6. Entity types must be EXACTLY as specified below (case-sensitive)
7. If entity type is Employee, extract the job `title`, `department`, and `team` if mentioned in the email body or in the email signature

### Allowed Entity Types by Document Type

**If `document_type` is `regular_email` then allowed entity types are:**
- Employee
- Company
- Department
- Team
- Task
- Project

**If `document_type` is `agenda` or `meeting_minute` then allowed entity types are all of the above plus:**
- Meeting

### Entity Type Definitions & Schemas

#### Employee
Individual people working at the company.

```json
{
  "type": "Employee",
  "name": "Full Name",
  "email": "email@company.com",
  "aliases": ["Full Name", "First Last", "Nickname", "F. Last"],
  "title": "Job Title",
  "department": "Department Name",
  "team": "Team Name"
}
```

#### Company
The organization itself or partner organizations.

```json
{
  "type": "Company",
  "name": "Company Name",
  "aliases": ["Company Name", "CompanyName Inc", "CN"]
}
```

#### Department
Organizational divisions (e.g., Flight Operations, Ground Services, Maintenance, Customer Service).

```json
{
  "type": "Department",
  "name": "Department Name",
  "aliases": ["Department Name", "Dept Name", "DEPT"]
}
```

#### Team
Working groups within departments.

```json
{
  "type": "Team",
  "name": "Team Name",
  "aliases": ["Team Name", "Team", "Group Name"],
  "department": "Parent Department Name"
}
```

#### Task
Specific work items and assignments.

```json
{
  "type": "Task",
  "name": "Task Name",
  "aliases": ["Task Name", "Task shorthand"],
  "description": "Task description",
  "status": "active OR completed",
  "assigned_to": ["Employee Name 1", "Employee Name 2"],
  "project": "Associated Project Name"
}
```

#### Project
Larger initiatives that encompass multiple tasks.

```json
{
  "type": "Project",
  "name": "Project Name",
  "aliases": ["Project Name", "Project Nickname", "Initiative Name"],
  "description": "Project description"
}
```

#### Meeting
Scheduled gatherings of employees (only for `agenda` or `meeting_minute` documents).

```json
{
  "type": "Meeting",
  "title": "Title of Meeting",
  "aliases": ["Meeting Title", "Alternative Title"],
  "date": "YYYY-MM-DD",
  "document_type": "agenda OR meeting_minute",
  "meeting_type": "direct_report OR team OR task OR executive OR project OR department"
}
```

**Meeting Type Classification:**
If the entity is of type `Meeting`, classify its `meeting_type` as ONE of the following based on the document content:
- `direct_report`: Meeting between an employee and their direct manager
- `team`: Meeting among members of the same team
- `task`: Meeting focused on a specific task
- `executive`: Meeting involving executives or high-level management
- `project`: Meeting focused on a specific project
- `department`: Meeting involving members of the same department

---

## Step 3: Relationship Extraction

### Relationship Extraction Rules

1. Extract ONE principal relationship (the most important/central relationship in the document)
{% block instructions_relationship %}
2. Extract multiple secondary relationships according to the criteria below
{% endblock %}
3. Relationship types must be EXACTLY as specified (case-sensitive)
4. **Temporality is REQUIRED** - must be one of: `start`, `end`, or `ongoing` (never null)
5. **Date Rules:**
   - If temporality is `start`: `beginning` must contain a valid date (YYYY-MM-DD), `end` is empty string
   - If temporality is `end`: `end` must contain a valid date (YYYY-MM-DD), `beginning` is empty string
   - If temporality is `ongoing`: both `beginning` and `end` are empty strings
6. Dates must be derived from the document - DO NOT fabricate dates
7. Extract `mentions`: direct quotes or paraphrases from the document that evidence this relationship

### Allowed Relationship Types by Document Type

**If the `document_type` is `regular_email` then the allowed relationship types are:**
- worksOn
- worksAt
- reportsTo
- memberOf
- departmentBelongsToCompany (secondary only)
- teamBelongsToDepartment (secondary only)

**If the `document_type` is `agenda` or `meeting_minute` then the allowed relationship types are all of the above plus:**
- organize
- attend
- mentionsTask
- mentionsProject
- mentionsEmployee
- mentionsTeam
- mentionsDepartment

### Relationship Type Definitions & Extraction Criteria

#### worksOn
**Pattern**: `Employee worksOn Task`

**Extract when:**
- The document explicitly states an employee is working on, assigned to, or responsible for a specific task
- Extract ONE relationship per task mentioned

**Example mentions:**
- "Sarah is working on the fleet maintenance audit"
- "John has been assigned the gate scheduling optimization task"

---

#### worksAt
**Pattern**: `Employee worksAt Company`

**Extract when:**
- The document mentions an employee works at a company
- The email sender uses a work email -> extract `sender worksAt company`
- The email recipient uses a work email -> extract `recipient worksAt company`

**Example mentions:**
- "As a flight coordinator at SkyWings Airlines..."
- Email from: john.smith@skywings.com -> John Smith worksAt SkyWings Airlines

---

#### reportsTo
**Pattern**: `Employee reportsTo Employee (manager)`

**Extract when:**
- The document explicitly mentions reporting structure
- Manager-employee language is used (e.g., "your progress on...", "I'd like you to...", "report back to me")

**Example mentions:**
- "As your manager, I wanted to check in on the crew scheduling..."
- "Please report your findings to Sarah"

---

#### memberOf
**Pattern**: `Employee memberOf Team` OR `Employee memberOf Department`

**Extract when:**
- The document signature contains the employee's team or department
- Collective language indicates team membership (e.g., "our team's progress", "we in the Ground Operations team")
- Explicit membership statements

**Example mentions:**
- Email signature: "John Smith | Ground Operations Team"
- "Our team has been coordinating baggage handling improvements..."

---

#### departmentBelongsToCompany
**Pattern**: `Department departmentBelongsToCompany Company`

**SECONDARY RELATIONSHIP ONLY** - never the principal relationship

**Extract when:**
- You have a principal relationship `Employee memberOf Department` AND the sender uses a work email
- Document signature contains department AND company information

**Example mentions:**
- Email signature: "Flight Operations Department | SkyWings Airlines"

---

#### teamBelongsToDepartment
**Pattern**: `Team teamBelongsToDepartment Department`

**SECONDARY RELATIONSHIP ONLY** - never the principal relationship

**Extract when:**
- The document explicitly states a team is within/part of a department

**Example mentions:**
- "The Crew Scheduling team within our Flight Operations Department..."

---

#### organize
**Pattern**: `Employee organize Meeting`

**Extract when:**
- Document type is `agenda` or `meeting_minute`
- The document identifies a meeting organizer/host

**Example mentions:**
- "Organized by: Sarah Johnson"
- "Meeting host: John Smith"
- "From: Sarah Johnson"

---

#### attend
**Pattern**: `Employee attend Meeting`

**Extract when:**
- Document type is `agenda` or `meeting_minute`
- Extract ONE relationship for EACH attendee listed

**Example mentions:**
- "Attendees: Sarah Johnson, Mike Chen, Emily Rodriguez"
- "To: Mike Chen, Emily Rodriguez"

---

#### mentionsTask
**Pattern**: `Meeting mentionsTask Task`

**Extract when:**
- Document type is `agenda` or `meeting_minute`
- A task is discussed or referenced in the meeting

**Example mentions:**
- "Agenda item 2: fleet maintenance audit task"

---

#### mentionsProject
**Pattern**: `Meeting mentionsProject Project`

**Extract when:**
- Document type is `agenda` or `meeting_minute`
- A project is discussed or referenced in the meeting

**Example mentions:**
- "Discussion of Fleet Modernization project timeline"

---

#### mentionsEmployee
**Pattern**: `Meeting mentionsEmployee Employee`

**Extract when:**
- Document type is `agenda` or `meeting_minute`
- An employee is mentioned in the meeting (who may not be an attendee)

**Example mentions:**
- "Sarah's progress on the safety audit will be reviewed"

---

#### mentionsTeam
**Pattern**: `Meeting mentionsTeam Team`

**Extract when:**
- Document type is `agenda` or `meeting_minute`
- A team is discussed or referenced in the meeting

**Example mentions:**
- "Ground Operations Team's efficiency metrics"

---

#### mentionsDepartment
**Pattern**: `Meeting mentionsDepartment Department`

**Extract when:**
- Document type is `agenda` or `meeting_minute`
- A department is discussed or referenced in the meeting

**Example mentions:**
- "Flight Operations Department's Q4 performance review"

---

## Relationship Schema

```json
{
  "type": "EXACT_TYPE_FROM_ABOVE",
  "source": "Source Entity Name",
  "target": "Target Entity Name",
  "context": "1-2 sentence context explaining this relationship from the document",
  "mentions": ["Direct quote 1 from document", "Direct quote 2 from document"],
  "temporality": "start OR end OR ongoing",
  "beginning": "YYYY-MM-DD or empty string",
  "end": "YYYY-MM-DD or empty string"
}
```

---

## Complete Output Schema

```json
{
  "document_type": "regular_email OR agenda OR meeting_minute",
  "entities": [
    {
      "type": "Entity Type",
      "name": "Entity Name",
      "aliases": ["alias1", "alias2"],
      ...additional fields based on entity type...
    }
  ],
  "relationships": [
    {
      "type": "Relationship Type",
      "source": "Source Entity Name",
      "target": "Target Entity Name",
      "context": "Context from document",
      "mentions": ["mention1", "mention2"],
      "temporality": "start OR end OR ongoing",
      "beginning": "YYYY-MM-DD or empty string",
      "end": "YYYY-MM-DD or empty string"
    }
  ]
}
```

---

## Extraction Checklist

Before finalizing your response, verify:

- [ ] Document type is correctly identified
- [ ] Exactly ONE principal relationship is extracted
- [ ] One or many secondary relationships are extracted as applicable
- [ ] Extract a Meeting entity for `agenda` or `meeting_minute` documents
- [ ] All extracted entities participate in at least one relationship
- [ ] All extracted relationships have a source and target entity present in the entities list
- [ ] All relationship types are spelled EXACTLY as specified
- [ ] All entity types are spelled EXACTLY as specified
- [ ] Every relationship has a valid temporality value (never null)
- [ ] Date fields follow rules (populated for start/end, empty for ongoing)
- [ ] All dates are in YYYY-MM-DD format
- [ ] All dates are extracted from the document (not fabricated)
- [ ] Mentions contain actual quotes/paraphrases from the document
- [ ] Aliases capture all variations found in the document
- [ ] Email addresses are extracted when available
- [ ] Title must be always extracted for Employee entities
- [ ] Output is valid JSON with no explanations or markdown

---

## EXAMPLES:

{{ few_shot_examples }}

---

## Your Task

Analyze the document below and return ONLY the JSON output following the schema. No explanations, no preambles, no markdown code blocks - just the raw JSON.

## DOCUMENT TO ANALYZE:
Filename: {{ document_filename }}
Content:
---
{{ document_content }}
---

RESPONSE:

\end{lstlisting}

\section{Topic Classification Prompt}
\label{app:topic-prompt}

The following is the extraction prompt used for the Topic Classification task.

\begin{lstlisting}

You are an expert topic classification system.

INSTRUCTIONS:
1. Return ONLY one of the provided labels, do not generate extra unnecessary text.
2. Do NOT generate an empty text as an answer, you ALWAYS MUST return one of the provided labels

# Few-shot examples:

## DOCUMENT TO ANALYZE:

Message-ID: d46bb20a-ef29-4936-bdc1-52cf20c9316a
From: Robert Lederer <Rlederer@poundfinancialgroupllc.com>
To: Nath Miller <miller.nath@gmail.com>
Date: 2024-03-01
Subject: Strengthening our checks and balances
MIME-Version: 1.0
Content-Type: text/plain; charset=UTF-8
Content-Transfer-Encoding: 7bit

Hey Nath, I wanted to touch base on something that's been on my radar lately. As we continue to focus on our broader business operations and how we're creating value across our portfolio, I think there's a real opportunity for us to tighten up our governance oversight. On that note, I'm newly working on subordination lien release verification, and I'm seeing firsthand how critical it is that we have solid verification processes in place.

You know, when you zoom out and look at portfolio management, a lot of it comes down to having strong accountability mechanisms that actually work. Right now, I'm thinking through some ways we could improve how we handle these verification workflows to make sure nothing slips through the cracks. It's not just about crossing boxes---it's about building confidence in our processes.

I think the key is making sure we've got clear ownership and tracking at every step. The current approach is solid, but there's definitely room to strengthen our controls and make the handoffs smoother between teams. Have you noticed any pain points on your end that we could address?

Would love to grab coffee or hop on a quick call this week to brainstorm some ideas. I'm pretty fired up about leveling up our accountability system, and I think your perspective would be really valuable here. Let me know what works for your schedule.

Thank you,
Robert

Robert Lederer
Analyst, Pound Financial Group LLC

P: +1 (415) 660-5966
E: Rlederer@poundfinancialgroupllc.com

## LABELS:

- Competitive Analysis
- Exit Planning
- Financial Analysis
- Legal Compliance
- Market Intelligence
- Market Research
- Network Development
- Non-business
- Operational Assessment
- Operational Improvement
- Opportunity Screening
- Portfolio Management
- Relationship Building
- Stakeholder Management
- Strategic Planning
- Technology Assessment
- Transaction Structuring

## RESPONSE:

Portfolio Management

....

# Your task:
Classify the following document using one of the labels provided (DO NOT REASON):

## DOCUMENT TO ANALYZE:

{{ document_content }}

## LABELS:

{{ topics }}

## RESPONSE:

\end{lstlisting}

\section{Topic Categories}
\label{app:topics}

\autoref{tab:topic-categories} shows the topic categories used for the topic classification task for Biocure (L) and Pound (XL) datasets.

\begin{table}[t!]
\centering
\small
\begin{tabular}{ll}
\toprule
\textbf{Biocure (L)} & \textbf{Pound (XL)} \\
\midrule
Non-business & Non-business \\
Approval Timeline Management & Exit Planning \\
Biomarker Identification & Financial Analysis \\
Clinical Data Analysis & Legal Compliance \\
Clinical Trial Planning & Market Intelligence \\
Competitive Intelligence & Market Research \\
Distribution Channel Management & Network Development \\
Efficacy Evaluation & Competitive Analysis \\
FDA Communication & Operational Assessment \\
Formulation Optimization & Operational Improvement \\
Healthcare Provider Education & Opportunity Screening \\
Intellectual Property Strategy & Portfolio Management \\
International Regulatory Coordination & Relationship Building \\
Key Opinion Leader Engagement & Stakeholder Management \\
Labeling Requirements & Strategic Planning \\
Manufacturing Compliance & Technology Assessment \\
Manufacturing Process Development & Transaction Structuring \\
Market Access Strategy & \\
Advisory Committee Preparation & \\
Partnership Collaborations & \\
Patient Assistance Programs & \\
Post Marketing Commitments & \\
Preclinical Research & \\
Pricing Negotiations & \\
Promotional Campaign Development & \\
Regulatory Strategy & \\
Regulatory Submission Preparation & \\
Safety Assessment & \\
Safety Reporting & \\
Sales Force Training & \\
Sales Performance Analytics & \\
\bottomrule
\end{tabular}
\caption{Topic Categories for Biocure (L) and Pound (XL).}
\label{tab:topic-categories}
\end{table}